\documentclass[a4paper,fleqn]{cas-sc} 

\usepackage[numbers]{natbib}

\usepackage{tikz}
\usepackage{latexsym}
\usepackage{float}
\restylefloat{table}
\usepackage{esvect}
\usepackage{amsmath}
\usepackage{amsthm}
\usepackage{setspace}
\usepackage{float}

\usepackage[toc,page]{appendix}
\newtheorem{proposition}{Proposition}
\newtheorem{lemma}{Lemma}

\usepackage[ruled,vlined]{algorithm2e}
\usepackage{array}
\newcolumntype{L}{>{\centering\arraybackslash}m{2.25cm}}

\begin{document}
\let\WriteBookmarks\relax
\def\floatpagepagefraction{1}
\def\textpagefraction{.001}

\shorttitle{Well-calibrated Confidence Measures for Multi-label Text Classification with a Large Number of Labels}

\shortauthors{Maltoudoglou et~al.}

\title [mode = title]{Well-calibrated Confidence Measures for Multi-label Text Classification with a Large Number of Labels}

\author[1,2]{Lysimachos Maltoudoglou}
\ead{lysimachosmalt@gmail.com}
\cormark[1]

\author[1,2]{Andreas Paisios}
\ead{paisiosa@gmail.com}

\address[1]{Machine Learning Research Group,  
Albourne Partners (Cyprus) Ltd  \\}

\address[2]{Computational Intelligence Research Lab.,
Frederick University, Cyprus}

\author[3,4]{Ladislav Lenc}[orcid=0000-0002-1066-7269]
\ead{llenc@kiv.zcu.cz}

\author[3,4]{Ji\v{r}\'{i} Mart\'{i}nek}
\ead{jimar@kiv.zcu.cz}

\author[3,4]{Pavel Kr\'{a}l}[orcid=0000-0002-3096-675X]
\ead{pkral@kiv.zcu.cz}

\address[3]{Dept. of Computer Science and Engineering, University of West Bohemia, 
Plze\v{n}, Czech Republic}

\address[4]{NTIS - New Technologies for the Information Society, University of West Bohemia, Plze\v{n}, Czech Republic}

\author[1,2]{Harris Papadopoulos}[orcid=0000-0002-6839-6940]
\ead{h.papadopoulos@frederick.ac.cy}
\cormark[1]

\cortext[1]{Corresponding author}

\begin{abstract}
We extend our previous work on Inductive Conformal Prediction (ICP) for multi-label text classification and present a novel approach for addressing the computational inefficiency of the Label Powerset (LP) ICP, arrising when dealing with a high number of unique labels. We present experimental results using the original and the proposed efficient LP-ICP on two English and one Czech language data-sets. Specifically, we apply the LP-ICP on three deep Artificial Neural Network (ANN) classifiers of two types: one based on contextualised (bert) and two on non-contextualised (word2vec) word-embeddings. In the LP-ICP setting we assign nonconformity scores to label-sets from which the corresponding p-values and prediction-sets are determined. Our approach deals with the increased computational burden of LP by eliminating from consideration a significant number of label-sets that will surely have p-values below the specified significance level. This reduces dramatically the computational complexity of the approach while fully respecting the standard CP guarantees. Our experimental results show that the  contextualised-based classifier surpasses the non-contextualised-based ones and obtains state-of-the-art performance for all data-sets examined. The good performance of the underlying classifiers is carried on to their ICP counterparts without any significant accuracy loss, but with the added benefits of ICP, i.e. the confidence information encapsulated in the prediction sets. We experimentally demonstrate that the resulting prediction sets can be tight enough to be practically useful even though the set of all possible label-sets contains more than $1e+16$ combinations. Additionally, the empirical error rates of the obtained prediction-sets confirm that our outputs are well-calibrated.
\end{abstract}

\begin{keywords}
text classification \sep multi-label \sep word2vec \sep bert \sep conformal prediction \sep label powerset \sep computational efficiency \sep nonconformity measure \sep confidence measure
\end{keywords}

\maketitle

        
\section{Introduction}

As the number of electronically produced texts continues to grow, automatic Text Classification (TC) becomes of increasing importance since it delivers an efficient approach to critical industry problems such as fake news detection, news categorisation, and sentiment analysis. The problem requires a system that can automatically assign a text, e.g. a document, to a label which is a set of zero, one or more classes. We can distinguish three cases: a binary scenario, where the text either belongs to a single class or not; a multi-class scenario, where the text belongs to exactly one out of many classes; or a multi-label scenario, where the text belongs to one or more out of a set of many classes. This work focuses on the latter because it frequently corresponds to real-world requirements and is often more challenging and resource demanding than its binary and multi-class counterparts.

Text classification, as a Natural Language Processing (NLP) task, is inextricably linked to text representation methods. Under this perspective, words are transformed into numerical vectors, i.e. Word Embeddings (WEs), which allow them to hold semantic information in a format that can be used in AI applications. Recent classification models build on different types of WEs like Word2Vec and contextualised embeddings which are produced from Artificial Neural Networks (ANNs). During the past decade, TC and various other NLP applications have seen a significant increase in scope, capabilities and performance gains \citep{Goth2016}. 

As the multi-label scenario requires a classifier to predict a set of classes (called \emph{label-set}), a first challenge involves formulating the problem in such a way that it allows for the training of the underlying machine-learning classification model so that it can produce meaningful results for this type of classification. Approaches are grouped into two broad categories: problem (or pattern) transformation methods and algorithm adaptation methods, as outlined in \cite{Tsoumakas-multi-label:}. In problem transformation methods the multi-label problem is reformulated, usually as a binary classification problem, as in \citep{lambr:brmlcp}, or as a multi-class classification problem using all relevant category combinations, i.e. power-set approach, as in \citep{boutell2004learning}. In the case of algorithm adaptation, solutions involve the modification of the underlying model in ways that allow it to handle multi-label outputs, e.g. in \citep{Zhang2007} by adapting the traditional \emph{k}-Nearest Neighbor algorithm, or in \citep{zhang2006multilabel} by adapting ANNs. Frequently however, problem transformation approaches are used in conjunction with algorithm adaptation, as in \citep{Madjarov2012}. 
An important limitation of most classification methods is that they do not provide any indication about the likelihood of their predictions being correct, and even in cases where they do provide probabilistic predictions their outputs can be misleading, as shown for instance in \cite{Lambrou_Papadopoulos2015} and \cite{papadopoulos2012}. Conformal Prediction (CP) aims to address this issue by supplementing conventional classification predictions with reliable confidence measures, which are guaranteed under the assumption of data exchangeability \citep{Shafer2007}. This confidence information can be of particular importance when dealing with classification tasks of low error tolerance, but can also be of use in a wider range of applications such as text-classification and, in general, in tasks where we are looking to set pre-defined specific confidence levels within which to operate. 

The extra confidence information provided over predictions comes with the cost of additional computational complexity. Various approaches to address this issue have been developed, such as Inductive Conformal Prediction (ICP) which also operates under the exchangeability assumption and provides the same guarantees as CP, but in a much more computationally efficient manner. However, in the multi-label classification scenario, when dealing with a large number of unique labels, the existing CP approaches either fail to decrease the computational load enough so as to be processed in a reasonable amount of time, or provide weaker CP guarantees, or practical inefficiencies as described in Section \ref{sec:cp_in_the_multi-label_setting}. 

Our previous work \cite{paisios2019deep} investigated the use of a Label Powerset (LP)-ICP combined with a text document classification model realized by a deep Convolutional Neural Network (CNN) with an additional trainable embedding layer. Due to its computational burden the approach was evaluated on a sub-set (20 most frequent classes) of the Reuters news-wire corpus~\citep{lewis2004rcv1} and results were compared against the performance of the underlying model as well as against the theoretical guarantees of CP. We also experimented on the use of different nonconformity measures and presented their effects on performance, particularly in the set-prediction mode. This paper extends the above, as follows:

\begin{itemize}

  \item 
  We attempt to address the computational limitations of LP-ICP on set-prediction evaluation by proposing a novel implementation of the approach that reduces the total number of computations (Section \ref{sec:efficient_lp_icp}). We mathematically establish the validity of the proposed approach and provide experimental results for multi-label text classification problems where it was previously computationally challenging.
  
  \item We obtain results for the complete Reuters 21578 corpus and also evaluate our proposed approaches on two additional benchmark corpora, namely: the "Czech text document corpus v 2.0" and the "Arxiv academic paper data-set" (Section \ref{sec:corpora}), using the efficient LP-ICP approach. For all data-sets we also produce ``reduced'' versions, based on the 20 most frequent classes, to apply the original ICP powerset approach for comparison purposes and for investigating the use of ICP in more detail. 

  \item Previously, our CNN-based classifier utilised only randomly initialised embeddings. However, as demonstrated in other studies (see Section \ref{sec:text_classification}), the use of pre-trained embeddings (specifically Word2Vec) can produce performance improvements, especially when fine-tuned for the specific text classification task. We therefore experiment with both configurations and compare the results. We also experiment with the recently proposed, state-of-the-art BERT-based model and compare it with the CNN-based models. We aim to provide well performing underlying classification models for CP and to evaluate the differences between the performance of contextualised (used in BERT) vs. non-contextualised embeddings (used in CNN). The classification results we obtained with the underlying models surpass the performance of published results from other studies.

\end{itemize}

This work is structured as follows: Section 2 presents an overview of the existing literature on text representation and text classification. Section 3 outlines the multi-label text classification model architectures used in our CP implementation. Section 4 gives an overview of the CP framework and its inductive version (ICP) and discusses the adaptation of the framework for multi-label problems. Section 5 presents the original LP-ICP approach and the proposed efficient LP-ICP and defines the nonconformity measures used in our experiments. In Section 6 we describe the data-sets used, our experimental set-up and the results of the experiments performed on the complete (efficient LP-ICP) and reduced data-sets (original LP-ICP), while Section 7 provides our conclusions and plans for future work.

\section{Related Work}
\label{sec:related_work}
This section summarises the related work in the NLP field, specifically the developments in text representation methods (Section \ref{sec:text_representation}) and their applications in the task of text classification (Section \ref{sec:text_classification}).  

\subsection{Text Representation}
\label{sec:text_representation} 

Most of the recent work on text representation is based on word embeddings, which are dense, real valued, high dimensional vector representations of text able to deliver semantic, similarity and analogy information between text entities. These have evolved out of earlier research in distributional semantics, primarily based on the distributional hypothesis, i.e. that "linguistic items with similar distributions have similar meanings" \citep{Firth1957}.
We can divide the existing WEs into two types: 1) static and 2) contextualised (dynamic), and their differences are summarised in Table~\ref{tab:Static_and_Contextual_Embeddings_Characteristics}.\par

\begin{table}[ht]
 \caption{Characteristics comparison of static and contextualized Word Embeddings}
 \label{tab:Static_and_Contextual_Embeddings_Characteristics}
    \centering
    \begin{tabular}{|| p{0.33cm} p{6.5cm}  p{6.5cm}||}
    \hline
    \multicolumn{3}{||c||}{Word Embeddings}\\
    & \quad \quad \quad \quad \quad Static  & \quad \quad \quad \quad \quad Contextualized \\
    \hline
    1 & A shallow window around each word is used. & The whole context (e.g. a sentence) is used. \\ [1ex]
    2 & Only specific word embeddings are generated - not a model that deals with NLP tasks. & They can produce both a trained model and the word embeddings.\\ [1ex]
    3 & They can't attribute different meanings to a single word depending on the context. & They are able to deal with polysemy.\\ [1ex]
    4 & They are usually discarded after the pre-processing. & They are the models themselves.\\ [1ex]
    \hline
    \end{tabular}
\end{table}

Static WEs are generated from shallow ANNs and are used as inputs to the first layer of the task-specific model. They are usually generated as pre-trained models, separately from the task-specific model, and a look-up table is required for their usage. 
Word2Vec \citep{mikolovEtAl2013}, is the first efficient static WE method that scaled up to deal with each word in a collective way and its superiority against its predecessors can be attributed to the significant reduction in its  computational complexity. 
Other representative examples of static WEs are the Global Vectors (GloVe) \citep{penningtonEtAl2014} and the FastText method \citep{Joulin2016}. The first combines Word2Vec with latent semantic analysis, attempting to address the weaknesses of both methods, while the latter can been seen as an extension of the Word2Vec where sub-word information is used to deal with the issue of absence of vector representations for out-of-vocabulary words.\par

Contextualised or dynamic WEs are produced from deep ANN structures and are able to capture more information than just the semantics of individual words. They have been developed based on the idea of transfer learning, the use of which has already demonstrated great performance improvements in various tasks within the field of computer vision. The concept of transfer learning is to develop models for tasks where data is abundant and then reuse them as the starting point for models of different but related tasks. 

Recently, contextualised WEs based on a deep ANN structure called the \emph{transformer} \citep{vaswani2017} were incorporated to models that successfully achieved new state-of-the-art performance on various NLP tasks,  especially on TC. Similar to the classical recurrent ANNs  (like LTSMs), the transformer was proposed to handle sequential data, commonly encountered in NLP tasks. The transformer, however, does not contain any component that handles the sequential nature of input data, but instead the positional encoding is utilised in order to inject some information about the relative or absolute position of the tokens in the sequence. In this way, transformers are able to process sequential data without being affected by the order in which data is placed within the sequence, e.g the final words of a sentence can be processed first. This characteristic paved the way for greater parallelisation than standard recurrent ANNs, resulting in training-time reductions. Moreover, it enabled training with much larger data-sets than what was previously possible.\par

Transformers are built on the Seq2Seq\footnote{Sequence to sequence models (Seq2Seq) \citep{sutskever2014} models proposed to address the problem of mapping input sequences to output sequences both of arbitrary lengths, highly encountered in machine translation. They are composed of an encoder and a decoder, similarly to auto-encoders. The encoder compresses the information of the input sequence (for instance a sentence of words), into a fixed-length vector, called the context vector, which is the output of the encoder and the input of the decoder. The output of the decoder is the final mapping of the input sequence.} model architecture and employ the attention mechanism\footnote{Attention \citep{bahdanau2014}, can be seen as a form of a fuzzy memory allowing the network to look at the input sequence, instead of forcing it to encode the whole input like in Seq2Seq models.} which was proposed for the NLP task of machine translation, and enabled neural machine translation models to outperform classic phrase-based ones. Most of the transformer-based models for NLP tasks make use of the transfer learning principle by being pre-trained as language models on large text corpora (e.g. Wikipedia). A limitation of transformer-based models is the huge number of trainable parameters, which is addressed by training them just once and then making them publicly available for further training on task-specific data-sets.\par 

The first application of transformers in NLP was the Generative Pre-Training model (GPT) \citep{Radfordetal2018} that achieved state-of-the-art performance in TC, as well as in 8 other NLP tasks. A later application of transformers, which made a more holistic use of the bidirectional information of sequential inputs, led to BERT \citep{devlinEtAl2019}, which obtained new state-of-the-art results on 11 NLP tasks, TC included. BERT has been especially influential for the next generation of transformer-based models, because of its efficiency, performance and ease of use.\par

\subsection{Text Classification}
\label{sec:text_classification}

The task of TC is nowadays mostly handled with ANNs and specifically with ``deep'' architectures that have a large number of layers and trainable parameters. A significant advantage of such networks is the ability to learn the features implicitly, without manual parameterisation and they have demonstrated significant performance gains in various NLP tasks, including: part-of-speech tagging, chunking, named-entity-recognition and semantic-role-labelling~\cite{collobert2011natural}. An alternative for the creation of efficient TC models is to exploit WEs and use them as features to train task-specific ANNs. In this case, the networks are usually ``shallow'' containing a small number of fully-connected layers. It is also possible to use the learned embeddings for the  initialisation of network parameters and allow for task-specific fine-tuning during the model training process~\citep{kim2014}.

Many proposed approaches have been based on CNNs. Examples include Zhang and LeCun~\cite{zhang2015text}, who use CNNs for ontology classification, sentiment analysis and single-label document classification. Their networks are composed of 9 layers out of which 6 are convolutional layers and 3 are fully-connected layers with different numbers of hidden units and frame sizes. They show that the proposed methods significantly outperform the baseline approaches (i.e. bag-of-words) on large English and Chinese corpora.

A well-known work by Kim~\cite{kim2014} uses pre-trained vectors from Word2Vec in the first layer (i.e. the look-up table). The author shows that the proposed models outperform the state-of-the-art in 4 out of 7 tasks, including sentiment analysis and question classification. In~\cite{conneau2016very}, a character based approach for TC is used, with 29 convolutional layers and the approach is evaluated on several tasks including sentiment analysis and news categorisation, reporting good results on larger corpora. 

In ~\cite{wang2018densely}, Wang et al. addressed the fact that most CNNs use filters of fixed lengths, which limits their ability to learn variable sized n-grams. They proposed a densely connected CNN, with multi-scale feature attention and showed that the network can learn meaningful text representations by selecting appropriate scales. The method outperformed most CNN-based approaches on several benchmark corpora. 

In~\cite{sun2019fine}, an approach for fine-tuning BERT for TC is presented and the method achieved state-of-the-art results on several classification and sentiment analysis data-sets. Huang et al.~\cite{huang2019hierarchical} presented an approach for hierarchical multi-label text classification with the use of text-category attention that aims at capturing the associations among texts and categories, and outputs the associated text-category representation. A method for document classification based on BERT was proposed in~\cite{adhikari2019docbert}. The authors have shown that a model based on BERT can achieve state-of-the-art results on TC tasks. They also utilise knowledge distillation from $BERT_{large}$ to a smaller LSTM, which reduces computational demands while preserving good accuracy. 
\section{Multi-label Text Classifiers Overview}
\label{sec:multi-label_text_classifiers_overview}

In this section, we describe the underlying classification models that were used as basis for our CP implementations. We compare three classifiers. The first two employ non-contextualised WEs and are denoted as ``randinit'' and ``word2vec''. Both use a CNN for the specific task of text classification. The ``randinit'' model uses a randomly initialized embedding layer and ``word2vec'' utilizes pre-trained Word2Vec WEs. Each model contains a trainable embedding layer, which is trained together with, and as a part of, the whole model. The third classifier, denoted as ``bert'', is built on the transformer-based BERT model that produces contextualised WEs. For all models we use binary cross-entropy~\citep{de2005tutorial} as a loss function, and the Adam~\citep{kingma2014} optimizer. For all experiments, we employ early stopping with patience of 3 epochs based on the validation F1-score.

\subsection{Classifiers with Non-Contextualised Embeddings}
\label{sec:classifier_with_non_contextualised_embeddings}
We use two kinds of non-contextualised embeddings for our CNN classifiers. The first relates to our previous work~\cite{paisios2019deep}, and it is a trainable embedding layer initialised randomly and trained as a part of the whole model. For comparison, we also utilise embeddings created by the Word2Vec method.  This is then fine-tuned during training to allow for adaptation on the task-specific target data.

\subsubsection{Text Representation}
\label{sec:text_representation_CNN}
The input text is first lower-cased and then tokenized in order to extract a sequence of all individual word and non-word tokens. Punctuation is removed and numbers are replaced by a special/reserved token. This is followed by vocabulary creation based on token frequencies from the training part of the corpus, i.e. we select the $|V|$ most frequent tokens. Words not present in the truncated vocabulary are replaced by an out-of-vocabulary (OOV) token (that it is itself part of the final vocabulary). The vocabulary then serves for the creation of the integer-encoded representations of documents (i.e. 1D integer vectors), where each token is represented by its vocabulary index.\par 

In order to be able to use the vectors as inputs to a neural-network we must ensure the same vector lengths. We set a~$WV_{max}$ as the maximum document length and documents exceeding this length are truncated. For shorter documents we use padding, achieved by filling the remaining length with a reserved ``padding'' token that also has a corresponding reserved integer counterpart.\par

The labels that belong to a given document must also be transformed to a suitable form. For this, we use a multi-hot encoding where the multi-hot representation is a vector of dimension $C$ equal to the number of categories present in the corpus. A zero/one encoding is used where a value of one indicates that the document belongs to the corresponding category and zero that it does not  (Figure  \ref{fig:multi-hot}).

\subsubsection{Classifier}
\label{sec:classifier_CNN}

\begin{figure}
    \centering
    \centerline{\includegraphics[width=9cm, angle=90]{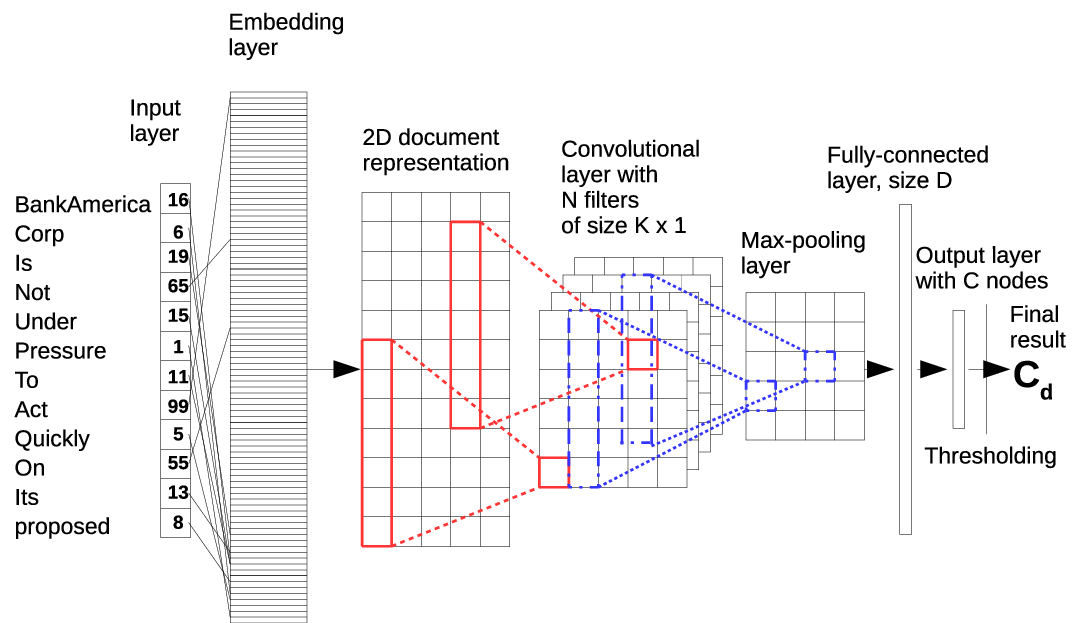}}
    \caption{Architecture of the CNN classifier used for ``randinit'' and ``word2vec'' models}
    \label{fig:cnn}
\end{figure}

Our classifier's architecture is depicted in Figure \ref{fig:cnn}. It was first used in~\cite{lenc2016} and it is a modified version of the CNN model proposed by Kim~\cite{kim2014}. Contrary to Kim's work, we use one-dimensional convolutional kernels, and our architecture uses one instead of three. It has been demonstrated that such an architecture is more suitable for longer text documents~\cite{lenc2017word}. 

The input layer receives integer-encoded word sequences of fixed length, described in the previous section. This is followed by an embedding layer which translates word indexes into real-valued vectors. In the case of the ``randinit'' model, this layer is initialised randomly, whereas the ``word2vec'' model uses the pre-trained Word2Vec vectors for the initialisation. The embedding layer produces 2D representations of the input texts, which are then fed to a convolutional layer consisting of 40 kernels of size $16\times 1$. Then, a max-pooling layer is used for dimensionality reduction, the output of which is flattened and fed to a fully-connected layer with 256 neurons.\par

A dropout with probability of 0.2 is applied after the pooling and fully-connected layers to achieve regularisation. The output layer size $C$ depends on the number of categories in the given data-set. We use a sigmoid activation function which is more suitable for the multi-label scenario we are dealing with. Finally, the values in the output layer are thresholded at 0.5 to obtain the classification result $C_d$ which is a set of predicted categories. For the whole training process we use a learning rate of $1e-3$.

\subsection{Classifier with Contextualised Embeddings}
\label{sec:classifier_with_contextualised_embeddings}
The utilised classifier and text pre-processing (tokenization) follows the work of \citep{devlinEtAl2019} and \citep{maltoudoglou2020} using the guidance provided by \cite{Howard2018} and \cite{sun2019fine} for the training process. We use the pre-trained model $BERT_{base}$ and add a fully connected output layer with sigmoid activations, for the task of text classification. We fine-tune the whole model (BERT and output layers) by adjusting the model parameters through re-training on new examples of in-domain text.

\subsubsection{Text Representation}
\label{sec:text_representation_bert}
BERT consists of $L$ identical sequential layers referred to as transformer blocks. Each one of these blocks is the encoder part of the general transformer structure 
and has hidden size $H$. This means that each block produces WEs of dimension $H$ which are the input of the next transformer block. The BERT structure comes in two flavours: $BERT_{base}$ with $L=12$ transformer layers, $H=768$ hidden size and $h=12$ attention-heads resulting in 110M total parameters and $BERT_{large} $ with $L=24, H=1024, h=16$ and 340M total parameters. Both versions of $BERT$ (base and large) are trained as a special language model called masked language model on the BookCorpus (800M words) and English Wikipedia (2500M words) with two objectives.  The first is to predict a word within a sequence of words regardless of the direction in which the sequence is processed; while the second is the next sentence prediction, where a binary classifier is used in order to determine whether two sentences come in sequence.

In this work we used the $BERT_{base}$ version\footnote{Downloaded from \url{https://github.com/google-research/bert}}, 
the input of which is a sequence of $T$ tokens, where $T \leq 512$ and the output is the representation of each token. BERT uses wordpiece tokens (e.g.``playing'' is split into tokens: ``play'' and ``\#\#ing'') with a 30k token vocabulary. The first token of every sequence is required to be the special classification token [CLS], while the token [SEP] is used between paired sentences and the token [PAD] is used for padding sequences of different lengths. The final output of the tokenization process is a sequence of integers corresponding to the vocabulary tokens. For words not found in the vocabulary the special token [UNK] is used.

\subsubsection{Classifier}
\label{sec:classifier_bert}
\begin{figure}
    \centering
    \includegraphics[width=\textwidth,keepaspectratio=true]{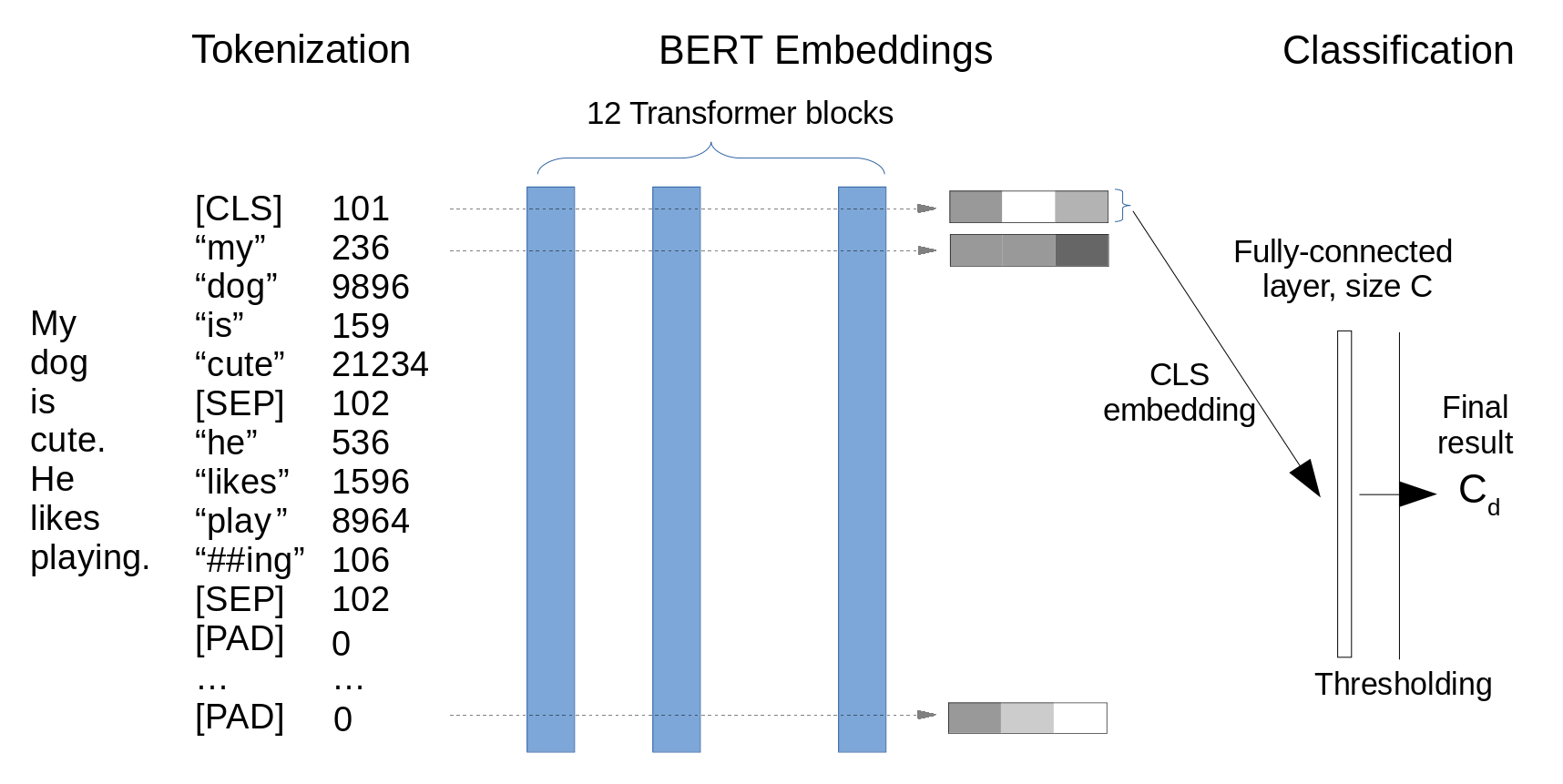}
    \caption{Architecture of BERT classifier}
    \label{fig:classifier}
\end{figure}

Our classifier is depicted in Figure \ref{fig:classifier}. Text is turned into tokens and then into the corresponding vocabulary indices which are the inputs of the first transformer block. Each transformer block outputs a sequence of $s$ real-valued vectors of length 768, where $s$ is the total number of tokens. WEs produced from lower transformer blocks capture more general information than the ones produced from the higher blocks. The WEs used in our model are the ones produced from the last (highest) transformer block (i.e. the $12^{th}$). 

The final hidden state of the [CLS], which is the first element of the output sequence of the $12^{th}$ transformer block, is the representation (WE) of the whole text and it is fed to the final layer of our classifier which is a fully-connected layer with input size 768 and output size equal to the total number of labels\footnote{We conducted preliminary experiments with 3 other set-ups for the text-specific layer. We experimented using: (i) 2 fully connected layers; (ii) 1 CNN layer; and (iii) 2 CNN layers. The performance difference was negligible, so we chose the 1 fully connected layer for simplicity as most bert implementations  for TC employ.}. To prevent over-fitting, we use dropout equal to $0.1$. The final output of the classifier is a vector of length equal to the number of labels with values in $[0,1]$, corresponding to the likelihood of each label being the true label. The final prediction $C_d$ is given by thresholding the classifier's predictions with a $0.5$ probability. \par

We address the danger of catastrophic forgetting \citep{Howard2018}, which eliminates the benefit of the information captured through the pre-training phase, using results from the work of \cite{sun2019fine}. The authors experimented with training strategies and concluded that an appropriate layer-wise decreasing learning rate and a preceding multi-task fine-tuning are beneficial for bert's overall performance. We only used the layer-wise decreasing learning rate for simplicity. Using this training strategy, we split the parameters $l$ in $\{l^{1}, \dots, l^{L} \}$ where $l^{i}$ is the learning rate for the parameters of the $i^{th}$ layer of BERT.  We set up the learning rate of the lowest layer $l^{1}$ and use $l^{k+1} =  \xi^{-1} l^{k}$, where $\xi \leq 1$ is a decay factor. When $\xi = 1$ all layers of BERT have the same learning rate. We set up $l^{1} = 2.5e-5$ and $\xi = 0.9$, as in \cite{sun2019fine}. For the parameters of the task-specific fully connected layer we used a much higher learning rate of $4e-4$ because of the random initialisation of its weights. \par
\section{Conformal Prediction}
\label{sec:conformal-prediction}

This section gives an overview of the CP framework starting with its original, transductive, version, followed by the computationally efficient inductive CP. Finally, it discusses its extensions to the multi-label classification setting and justifies the choice of approaches followed in this work.

\subsection{Transductive-CP}
\label{sec:transductive-cp}

The objective in a typical classification problem is to correctly predict the category to which a new, unseen example belongs. This is achieved by training a classification model on a set of known examples, i.e. training instances, of the form $\{(x_1, y_1),\dots,(x_n, y_n)\}$, where each pair $(x_i, y_i)$ consists of a set of attributes $x_i \in \mathbb{R}^{s}$ and a corresponding category $y_i \in \{Y_1,...,Y_c\}$, i.e. its corresponding classification. 
A new example $x_{n+1}$ is then assigned to a classification $Y_j$, which is singled out from the full set of possible classifications $\{Y_1,...,Y_c\}$, usually based on post-processing of the resulting probabilistic (or non-probabilistic) model outputs. 

In the CP framework, and provided that the assumption of exchangeability holds for $(x_i, y_i), i = 1,\dots,n+1$, the objective is to produce a \emph{set-prediction} $\Gamma_{n+1}^\epsilon \subseteq \{Y_1,...,Y_c\}$ that will contain the true category of the example $x_{n+1}$ with a probability $1 - \epsilon$, where $\epsilon$ is the \emph{significance-level}, \citep{Vovk2005}. This is achieved by first assuming all possible classifications $Y_j \in \{Y_1,...,Y_c\}$ for $x_{n+1}$ and then determining the likelihood that each of the extended sets:
\begin{equation}
\label{eq:extended_set}
    \{(x_1,y_1),...,(x_n,y_n),(x_{n+1},Y_j)\},
\end{equation}
is exchangeable. Since $(x_{n+1},Y_j)$ is the only artificial pair, we are in effect assessing the likelihood of $Y_j$ being the true category of $x_{n+1}$. This likelihood is quantified by first mapping each pair $(x_i,y_i)$ in (\ref{eq:extended_set}) to a numerical score using a \emph{nonconformity-measure}, $A$:
\begin{subequations}\label{eq:nmdef}
\begin{align}
\alpha^{Y_j}_i &= A(\{(x_1, y_1), \dots, (x_n, y_n), (x_{n+1}, Y_j)\}, (x_i,y_i)), \mspace{15 mu} i = 1, \dots, n, \\
\alpha^{Y_j}_{n+1} &= A(\{(x_1, y_1), \dots, (x_n, y_n), (x_{n+1}, Y_j)\}, (x_{n+1}, Y_j)).
\end{align}
\end{subequations}
The score $\alpha^{Y_j}_i$, called the \emph{nonconformity score} of instance $i$, indicates how nonconforming, or strange, it is for $i$ to belong in~(\ref{eq:extended_set}). In effect, the nonconformity measure is based on a conventional machine learning algorithm, called the \emph{underlying algorithm} of the corresponding CP, and measures the degree of disagreement between the actual label $y_i$ and the prediction $\widehat{y_i}$ of the underlying algorithm, after being trained on~(\ref{eq:extended_set}). 

The nonconformity score $\alpha^{Y_j}_{n+1}$ is then compared to the nonconformity scores of all other examples to find out how unusual $(x_{n+1}, Y_j)$ is according to the nonconformity measure used. This comparison is performed with the function
\begin{equation}
\label{pvalue}
  p(Y_j) = \frac{|\{i = 1, \dots, n : \alpha^{Y_j}_i \geq \alpha^{Y_j}_{n+1}\}|+1}{n+1},
\end{equation} 
the output of which is called the \emph{p-value} of $Y_j$. 
An important property of (\ref{pvalue}) 
is that $\forall \epsilon \in [0, 1]$ and for all probability distributions $P$ on $\mathcal{Z}$,
\begin{equation}
\label{eq:validity}
  P^{n+1}\{((x_1,y_1), \dots, (x_n, y_n), (x_{n+1},y_{n+1})):p(y_{n+1}) \leq \epsilon\}\leq \epsilon;
\end{equation}
a proof of which can be found in \citep{Vovk2005}. According to this property, if $p(Y_j)$ is under some 
very low threshold, say $0.05$, this means that $Y_j$ is highly unlikely 
as the probability of such an event is at most $5\%$ if~(\ref{eq:extended_set})
is exchangeable. Therefore we can reject it and have at most $\epsilon$ chance of being wrong. Consequently, the output of CP for a predefined confidence level $1 - \epsilon$ is the \emph{prediction set}: 
\begin{equation}
\label{eq:prediction_set}
\Gamma^\epsilon_{n+1} = \{ Y_j : p(Y_j) > \epsilon \},
\end{equation}
which according to property (\ref{eq:validity}) will not contain the true label of the instance with at most $\epsilon$ probability.

Alternatively, CP can produce a single prediction, called \emph{forced prediction}, where the highest p-value classification is selected and accompanied by a confidence score equal to one minus the second highest p-value and a credibility score equal to the p-value of the predicted classification (i.e. the highest p-value). Confidence is an indication of how likely the prediction is of being correct compared to all other possible classifications, while credibility indicates how suitable the training set is for the particular instance; specifically, a very low credibility value indicates that the particular instance does not seem to belong to any of the possible classifications. 

\subsection{Inductive-CP}
\label{sec:inductive-cp}
Suppose we want to apply CP to a set of $g$ instances with $c$ possible classifications. In the original \emph{transductive} setting this is achieved by re-training the underlying classification model $g \times c$ times, i.e. per instance and for each possible classification, in order to obtain the necessary p-values. This process can become computationally inefficient and usually prohibitive, particularly for data-sets with a high number of instances and for resource-demanding underlying models (such as deep ANNs).

An alternative to the above is \emph{Inductive} Conformal Prediction (ICP), first proposed in \cite{papa:icm-rr} and \cite{icp} for regression and classification tasks, 
respectively. In the ICP setting, the underlying model is trained only once thus generating a single general ``rule'' which is then applied on each test instance. 

Specifically, the training data-set is first split into two components, the \emph{proper-training set} $\{(x_1,y_1),\dots, (x_q,y_q)\}$ and the \emph{calibration set} $\{(x_{q+1},y_{q+1}), \dots, (x_n,y_n)\}$. The underlying model is trained on the proper-training set and the resulting model is used for calculating the nonconformity scores of the calibration instances as:
\begin{equation}
\label{eq:calibration_non_nonformity_scores}
    \alpha_i = A(\{(x_1, y_1), \dots, (x_q, y_q)\}, (x_i,y_i)), \mspace{15 mu} i = q+1, \dots, n.
\end{equation}
The trained model and the nonconformity scores of the calibration set form the general ``rule'' of the ICP. Then, the nonconformity score for each assumed classification $Y_j$ of every test instance $x_{n+m}, m = 1, \dots, g$, is calculated as:
\begin{equation}
\label{eq:label_set_non_conformity_scores}
    \alpha^{Y_j}_{n+m} = A(\{(x_1, y_1), \dots, (x_q, y_q)\}, (x_{n+m},Y_j)),
\end{equation}
and is used together with the nonconformity scores of the calibration instances to calculate the p-value:
\begin{equation}
\label{eq:pvalueicp}
  p(Y_j) = \frac{|\{i = q+1, \dots, n : \alpha_i \geq \alpha^{Y_j}_{n+m}\}|+ 1}{n-q+1}.
\end{equation} 

The rest of the process is the same as described for the transductive version. 

\subsection{CP in the Multi-label Setting}
\label{sec:cp_in_the_multi-label_setting}
Unlike the multi-class classification setting, in multi-label classification each instance can belong to multiple classes, e.g. a news-wire might be tagged under both \emph{politics} and \emph{economy}. Therefore the true classification (\emph{label-set}) $y_{n+m}$ of each instance $x_{n+m}$ is a set $Y_j \in \mathcal{P}(\{\Psi_1, \dots, \Psi_d\})$ \footnote{The powerset of a set A, i.e. $\mathcal{P}(A)$, is a set that contains all subsets of $A$ including $\emptyset$ and $A$ itself.}, where $\Psi_1, \dots, \Psi_d$ are the unique classes.

Different variations of CP have been proposed for handling multi-label classification problems through problem reformulation. These include decomposing each multi-label instance into a number of single-label examples (\emph{Instance Reproduction} (IR)) \citep{wang:Instance_Reproduction2014}; or decomposing the problem into several independent binary classification problems - one for each unique class (\emph{Binary-Relevance} (BR)) \citep{lambr:brmlcp, wang:mlcp}; or transforming the problem into a multi-class one by treating each label-set as a unique class (\emph{Label Powerset} (LP)) \citep{papa:mlcp}.

As the standard guarantee provided by CP is that $P(y_{n+m} \notin \Gamma^\epsilon_{x_{n+m}}) \leq \epsilon$ and since in the multi-label setting $y_{n+m}$ is a label-set, the natural extension of CP to multi-label classification is for the prediction set $\Gamma^\epsilon_{x_{n+m}}$ to correspond to a set of label-sets, i.e. $\Gamma^\epsilon_{x_{n+m}} \subseteq \mathcal{P}(\{\Psi_1, \dots, \Psi_d\})$, with at most $\epsilon$ chance of not containing the true label-set $y_{n+m}$. This is the guarantee provided by the approaches proposed in \citep{lambr:brmlcp} and \citep{papa:mlcp}. The other variations on the other hand provide somewhat weaker guarantees. Specifically, the approaches proposed in \citep{wang:Instance_Reproduction2014} and \citep{wang:mlcp} provide a prediction set $\Gamma^\epsilon_{x_{n+m}} \in \mathcal{P}(\{\Psi_1, \dots, \Psi_d\})$ with the guarantee that the chance of it not containing any element of $y_{n+m}$ is at most $\epsilon$, i.e. $ \forall \Psi_i \in y_{n+m} \implies P( \Psi_i \notin \Gamma^\epsilon_{x_{n+m}}) \leq \epsilon$. This does not include any bound on the chance of $\Gamma^\epsilon_{x_{n+m}}$ containing labels that are not in $y_{n+m}$.

In this work we follow the LP approach proposed in \citep{papa:mlcp} for three main reasons. Firstly, we want our approach to provide equally strong guarantees as those provided by CP in other settings. Secondly, the BR approach proposed in \citep{lambr:brmlcp} applies CP independently for each label and then combines the resulting prediction sets. In order to produce valid prediction sets for a significance level $\epsilon$ without making any extra assumptions, the significance level of the individual CPs should be set to $\epsilon/d$ (where $d$ is the number of labels) as derived from the Bonferroni general inequality. This is highly impractical for tasks with a high number of labels, such as the ones we are dealing with in this work as it would result in extremely low significance levels. Thirdly, the LP transformation has the added advantage of enabling the development of more accurate nonconformity measures by taking label correlations into account, as opposed to dealing with each label individually. The main disadvantage of LP is its computational complexity with respect to the number of labels, which grows exponentially. For this reason, we follow the inductive version of the CP framework and propose a modification of the original approach that provides significant gains in terms of computational efficiency while producing exactly the same outputs.

\section{Label Powerset ICP}
\label{sec:label_powerset_icp}

In the LP-ICP setting, all possible \emph{label-sets} $Y_j \in \{Y_1, \dots, Y_c\} \subseteq \mathcal{P}(\{\Psi_1, \dots, \Psi_d\})$ are treated as candidate classifications and are assigned p-values. For a data-set with $d$ single labels we denote the total number of possible label-sets as $c(d) = |\mathcal{P}(\{\Psi_1, \dots, \Psi_d\}| = \sum^{d}_{k=1}\binom{d}{k} = \frac{d!}{1!(d-1)!} +\dots + 1$. Consequently, $c$ gets extremely large, even for relatively small $d$ and therefore we consider only class combinations up to the maximum observed label cardinality $l, l \leq d$ of the proper training-set. In this way, we deal with only $c(d,l) = \sum^{l}_{k=1}\binom{d}{k} = \frac{d!}{1!(d-1)!} + \dots + \frac{d!}{l!(d-l)!}$ possible label-sets by making the assumption that the true label-set of any test instance contains at most $l$ labels.

However, for the data-sets used in this study $d \geq 54$ and $l \geq 8$, which still lead to prohibitively large $c(d,l)$. For example, for the complete Reuters 21578 data-set where $d=90$ and $l=15$, we need to consider and evaluate label-sets of number $c(90,15) = \sum^{15}_{k=1}\binom{90}{k} = \frac{90!}{1!(90-1)!} +\dots + \frac{90!}{15!(90-15)!} >> 1e+16$. We deal with computational complexity in 2 ways, leading to 2 different sets of experimental results: (i) by considering the reduced data-sets with the $d = 20$ most frequent classes and applying the original LP-ICP approach; (ii) by applying a novel approach of an efficient version of LP-ICP, proposed in this study on the full data-sets. This section presents both the original and efficient LP-ICPs as well as the nonconformity measures employed. 
\subsection{Original Label Powerset ICP}
\label{sec:original_label_powerset_icp}
Assuming the general ICP framework (Section \ref{sec:inductive-cp}), the original LP-ICP approach is shown in Algorithm \ref{alg:lp-icp}.

\begin{algorithm}
\label{alg:lp-icp}
\caption{LP ICP}
\SetAlgoLined
\DontPrintSemicolon
\KwIn{Calibration set $\{(x_{q+1},y_{q+1}), \dots, (x_n,y_n)\}$,  classifier's raw predictions (prior to thresholding) $o(x_i), i = q+1,\dots,n$ and $o(x_{n+m})$, nonconformity measure $A$ and significance level $\epsilon$} 
\KwOut{Prediction set $\Gamma^\epsilon_{n+m}$}

\nl Calculate nonconformity scores for calibration set $\alpha_i, i = q+1, \dots, n$ using (\ref{eq:calibration_non_nonformity_scores}), i.e. $\alpha_i = A(o(x_i), y_i)$ \;

\nl Calculate nonconformity scores $\alpha^{Y_j}_{n+m}$ for all possible label-sets $Y_j \in \{Y_1, \dots, Y_c\}$ using (\ref{eq:label_set_non_conformity_scores}), i.e. $\alpha^{Y_j}_{n+m} = A(o(x_{n+m}), Y_j)$\;

\nl Get p-values $p(Y_j)$ for each label-set $Y_j$ using (\ref{eq:pvalueicp})

\nl Obtain prediction set $\Gamma^\epsilon_{n+m} = \{ Y_j : p(Y_j) > \epsilon \}$\;
\end{algorithm}

\subsection{Efficient LP-ICP}
\label{sec:efficient_lp_icp}
In this section, we outline an approach for efficiently obtaining LP-ICP results for data-sets with large $d$, in an attempt to further address the issue. Specifically, our approach aims to decrease the computational cost of the ICP process by not considering, and consequently not evaluating, label-sets for which we can determine in advance that they are not going to be included in the prediction sets up to the pre-specified significance level $\epsilon$. This a priory knowledge is derived from $\epsilon$ and the nonconformity scores of the calibration set. To the best of our knowledge no studies have been published that address the difficulty of implementing ICP classification problems in this way. However, relevant studies for regression problems have been published, as in~\cite{papa:icm-rr} and~\cite{icp-regression}, from which this work has been inspired.

We reformulate the LP-ICP approach described in Section \ref{sec:original_label_powerset_icp} in which the prediction set for test instance $x_{n+m}$ is defined as $\Gamma_{n+m}^\epsilon = \{Y_j : p(Y_j) > \epsilon \}$. Using (\ref{eq:pvalueicp}) for a pre-specified significance level $\epsilon$ we get:
\begin{equation}
\label{eq:problem_reformolation}
\begin{split}
    \Gamma_{n+m}^{\epsilon} = \{Y_j : \frac{|\{i = q+1, \dots, n : \alpha_i \geq \alpha^{Y_j}_{n+m}\}|+ 1}{n-q+1} > \epsilon \} \mspace{40 mu} 
     \Leftrightarrow \mspace{12 mu} \\
    \Gamma_{n+m}^{\epsilon} = \{Y_j : |\{i = q+1, \dots, n : \alpha_i \geq \alpha^{Y_j}_{n+m}\}| > \epsilon(n-q+1) -1 \}
\end{split}
\end{equation}

We distinguish all the nonconformity scores  $\alpha_k \in \{\alpha_i, i = q+1,\dots,n \}$ that (in place of $\alpha^{Y_j}_{n+m}$) satisfy (\ref{eq:problem_reformolation}) and collect them in set $T$:
\begin{equation}
\label{eq:set_T_equation}
T  = \{\alpha_k, k = q+1,\dots,n  : |\{i = q+1, \dots, n : \alpha_i \geq \alpha_k\}| > \epsilon(n-q+1) -1 \}
\end{equation}
Then $\alpha_{i_0} = \alpha_{i_0}(\epsilon)  = \max (T)$ is a threshold for all $\alpha^{Y_j}_{n+m}$ that if surpassed, label-set $Y_j$ is not included in $ \Gamma_{n+m}^{\epsilon}$, that is:
\begin{equation}
\label{eq:conclution_for_power_set_ICP}
    \alpha^{Y_j}_{n+m} > \alpha_{i_0} \Leftrightarrow Y_j \notin \Gamma_{n+m}^{\epsilon}
\end{equation}

To make this clearer, consider the example of having $n-q = 999$ different calibration set nonconformity scores from 1 to 999, i.e. $\alpha_i = i, i = 1,\dots,999 $ and $\epsilon = 0.05$. Then $T = \{1,\dots,950\}$, $max(T) = \alpha_{i_0}(0.05) = 950$ and $\forall  \alpha^{Y_j}_{n+m} > 950 \Leftrightarrow Y_j \notin \Gamma_{n+m}^{0.05}$.

Under this problem formulation, our goal for reducing the computational complexity of instance $x_{n+m}$ becomes finding a set $Q_{n+m} : \Gamma_{n+m}^{\epsilon} \subseteq Q_{n+m} \subseteq \{Y_1, \dots, Y_c\}$ that is as small as possible. Therefore $\Gamma_{n+m}^{\epsilon} = \{Y_j \in Q_{n+m}: \alpha^{Y_j}_{n+m} \leq \alpha_{i_0}\}$, as opposed to considering all possible label-sets. Given $Q_{n+m}$ with cardinality $|Q_{n+m}| = c_{n+m}^q$, the algorithm for the efficient LP-ICP is the original (Algorithm \ref{alg:lp-icp}), but with $Y_j \in Q_{n+m}$, instead of $Y_j \in \{Y_1, \dots, Y_c\}$ in step 2. Since the computational load is directly linked to the total number of class combinations considered, this improvement is significant as it narrows down the total amount of computations significantly when $c_{n+m}^q \ll c$. In the case where $c_{n+m}^q = c$, this corresponds to the original LP-ICP approach. 

\subsubsection{Efficient LP-ICP Based on Minimum Nonconformity Change}
\label{sec:efficient_lp_icp_based_on_max_min_nonconformity_mesurment}
This section describes a way to construct the set $Q_{n+m}$ that possesses the property of $\Gamma_{n+m}^{\epsilon} \subseteq Q_{n+m}$ based on information provided from (\ref{eq:conclution_for_power_set_ICP}) therefore $Q_{n+m} = Q_{n+m}(\alpha_{i_0}(\epsilon))$. It is worth mentioning that this approach results in a different set $Q_{n+m}$ for each test instance $x_{n+m}$ . This means that, in general, different number of computations are required per instance, as opposed to the original LP-ICP. Our approach is valid under a broad family of nonconformity measures based on metrics that are often used by practitioners, i.e. the norms $L_p, p \geq 1$ (see Section \ref{sec:nonconf}). Set $Q$ contains all label-sets that differ up to $t_{n+m}=t_{n+m}(\alpha_{i_0}(\epsilon))$  labels from the label-set with the lowest nonconformity score which is the $z_{n+m}$. i.e. the underlying classifier's prediction using the threshold 0.5. Thus, set $Q_{n+m}$ is defined as:
\begin{equation}
    \label{eq:definition_of_q}
    Q(t_{n+m}) = \{Y_j: |Y_j  \triangle z_{n+m}| < t_{n+m}\}, 
\end{equation}
where $\triangle$ is the symmetric difference operator\footnote{The symmetric difference of sets $A$ and $B$ is the union without the intersection, i.e. $A \triangle B = (A - B) \cup (B - A)$} and $t_{n+m} \in \{1,\dots,d\}$. Our implementation of efficient LP-ICP is given in Algorithm \ref{alg:find_t}. The intuition for our approach and a proof that $\Gamma_{n+m}^{\epsilon} \subseteq Q(t_{n+m})$ is given in the Appendix, Section  \ref{construct_set_q_based_on_minimum_nonconformity_change} and Proposition \ref{proposition_1}, respectively.

\begin{algorithm}
\label{alg:find_t}
\caption{Efficient LP-ICP based on Minimum Nonconformity Change}
\SetAlgoLined
\DontPrintSemicolon
\KwIn{Calibration set $\{(x_{q+1},y_{q+1}), \dots, (x_n,y_n)\}$, classifier's raw predictions (prior to thresholding) $o(x_i), i = q+1,\dots,n$ and $o(x_{n+m})$,  nonconformity measure $A \in L_p, p \geq 1$ and significance level $\epsilon$} 
\KwOut{Prediction set $\Gamma^{\epsilon}_{n+m}$ }

\nl Calculate nonconformity scores for calibration set $\alpha_i, i = q+1, \dots, n$ using (\ref{eq:calibration_non_nonformity_scores}), i.e. $\alpha_i = A(o(x_i), y_i)$ \;

\nl Calculate nonconformity threshold $\alpha_{i_0}$, i.e. $\alpha_{i_0}(\epsilon) = max(T)$,  where $T$ is found using (\ref{eq:set_T_equation})\;

\nl Initialise label-set $Y_{current} \gets z_{n+m}$ ($\alpha^{z_{n+m}}_{n+m} = \min\limits_{j = 1, \dots, c}(\alpha^{Y_j}_{n+m}) $, see Lemma \ref{lemma_1})\;

\nl Set up index function $s(t), t = 1,\dots, d$ for sorting labels $\Psi_k,k =1,\dots,d$ in ascending order according to how much they affect the nonconformity score $\alpha^{z_{n+m}}_{n+m}$ if altered, i.e. $s(1)$ is the index of the label with raw prediction closer to $0.5$

\nl $t_{n+m} \gets 0$  \;
\While{$\alpha^{Y_{current}}_{n+m} \leq \alpha_{i_0}$}
    {   $t_{n+m} \gets t_{n+m} + 1$ \;
        \eIf{$\Psi_{s(t_{n+m})} \in Y_{current}$}
            {$Y_{current} \gets Y_{current} -\{ \Psi_{s(t_{n+m})} \}$} 
            {$Y_{current} \gets Y_{current} \cup \{ \Psi_{s(t_{n+m})} \}$}
        }

\nl Construct set $Q(t_{n+m})$ using (\ref{eq:definition_of_q})

\nl Follow steps of algorithm \ref{alg:lp-icp} using label-sets $Y_j$ from set $Q(t_{n+m})$ instead of $\{Y_1, \dots, Y_c\}$ from step 2 onwards 

\end{algorithm}

\subsection{Nonconformity Measures} 
\label{sec:nonconf}
In order to produce valid targets for our classifiers, the label-set of each training/test instance is  transformed into a multi-hot  representation. Specifically, we associate each instance $(x_i, y_i)$, where $y_i \in  \mathcal{P}(\{\Psi_1,\dots,\Psi_d\})$, with a 1D binary label-vector, i.e. $\vv{y}_i =  (t_i^1, \dots, t_i^d)$, where $t_i^j=1$ iff $\Psi_j \in y_i$ and $t_i^j=0$ otherwise. An example is shown in Figure \ref{fig:multi-hot}.\par 

\begin{figure}[h!]
    
    \begin{center}
    \begin{tikzpicture}
        [box/.style={rectangle,draw=black, ultra thick, minimum size=1cm}]
        \foreach \x/\y in {0/0, 1/0,2/1,3/0,4/0,5/0,6/0,7/1,8/0,9/0,10/0,11/0,12/0,13/0,14/0}
        \node[box] at ( \x,0){\y};
    \end{tikzpicture}
\end{center}
\caption{Multi-hot representation for the label-set $Y_j = \{\Psi_3,\Psi_8\}$, for $d = 15$}
\label{fig:multi-hot}
\end{figure}

We compute nonconformity scores $\alpha_i$ and $\alpha^{\vv{Y}_j}_{n+m}$ for the raw predictions of calibration-set instances $x_i, i = q+1,\dots,n$ and test-set instances $x_{n+m}$, respectively, as described in Algorithm \ref{alg:lp-icp}. In this work we focus on frequently used metrics, i.e. the $L_p$ norms or $\| \|_{p}$ \footnote{
$   \forall \vv{e} \in \mathbb{R}^d \text{with} \vv{e}  = (e_1, \dots , e_d), 
    L_p ( \vv{e})  = \| \vv{e} \|_{p} = \sqrt[p] {\sum _{k=1}^{d} | e_{k} | ^p }, p \geq 1.
$
}. Factor $p$ is used for controlling the sensitivity of the nonconformity measure, since larger values of $p$ focus on the label with the highest prediction error, while smaller values take into account errors in all labels equally. To better understand the behaviour and the impact of factor $p$ in ICP we employ 3 different nonconformity measures in our experiments with for $p = 2,4$ and $8$. Specifically our nonconformity measures are:
\begin{subequations}
    \label{eq:nonconformity1}
    \begin{align}
    \alpha_i = \| \vv{o}_i - \vv{y}_i \|_{p}  \mspace{30 mu}
    \\
    \alpha^{Y_j}_{n+m} =\| \vv{o}_{n+m} - \vv{Y}_j \|_{p}
    \end{align}
\end{subequations}
where $\vv{o}_i, \vv{o}_{n+m} \in  R^{d}$ are the raw predictions of the classifier for calibration and test set instances, respectively. The true label-set of the $i^{th}$ instance of the calibration set is denoted as $y_i$.

\section{Experimental Results}
\label{sec:experimental-results}
This section describes our experimental set up and presents classification results: (i) without the use of ICP; (ii) with the use of the original LP-ICP approach on the reduced data-sets; and (iii) with the use of the efficient LP-ICP approach on full data-sets. Our hardware setup is a Ryzen Threadripper 2950X 3.5GHz 16-Core,  GeForce RTX 2080 Ti 11GB and 64 GB DDR4.

\subsection{Corpora}
\label{sec:corpora}

We present experimental results for three data-sets, two consisting of English-language documents and one consisting of Czech-language documents, the statistics of which are summarised in Table~\ref{tab:corpora_stats}.
The ``word2vec'' model uses WEs trained on Google News\footnote{\url{https://s3.amazonaws.com/dl4j-distribution/GoogleNews-vectors-negative300.bin.gz}} and on the Czech Wikipedia. The Bert classifier uses $BERT_{base}$ as described in Section \ref{sec:text_representation_bert} and  SlavicBERT\footnote{https://huggingface.co/DeepPavlov/bert-base-bg-cs-pl-ru-cased} \citep{arkhipov2019} which was initializated from multilingual $BERT_{Base}$ and later trained  on Russian News and four Wikipedia versions: Bulgarian, Czech, Polish, and Russian.

For our experiments with ICP, we  additionally partition the training set into proper training and calibration sets. The calibration documents are selected randomly from the training set in a stratified manner  (with respect to label distributions). The size of the calibration data-set is 999 in all cases apart from AAPD and AAPD20 where it is 4999 (because of its large size). The data-sets CTDC20, Reuters20 and AAPD20 are reduced versions of the full ones, generated by selecting the documents that belong to the 20 most frequent classes in all cases. 

\begin{table}[ht]
\centering
\footnotesize{
  \caption{Corpora statistics; \textit{Docs} refers to the total number of documents in the corpus.}
  \label{tab:corpora_stats}
  \begin{tabular}{c|cccccccc}
    \textbf{} & \textbf{Docs} & \textbf{Proper train} & \textbf{Calibration} &\textbf{Test} & \textbf{Validation} & \textbf{Labels} & \textbf{avg. labels} & \textbf{avg. words} \\
    \hline
    \textbf{CTDC} & 14,690 & 8,565 & 999 & 2,391 & 2,735 & 60 & 2.64 & 275 \\
    \textbf{CTDC20} & 13,702 & 8,032 & 999 & 2,274 & 2,397 & 20 & 1.96 & 269 \\
    \hline
    \textbf{Reuters} & 10,788 & 5,770 & 999 & 3,019 & 1,000 & 90 & 1.24 & 165 \\
    \textbf{Reuters20} & 9,855 & 5,118 & 999 & 2,738 & 1000 & 20 & 1.16 & 163 \\
    \hline
    \textbf{AAPD} & 55,840 & 48,841 & 4,999 & 1,000 & 1,000 & 54 & 2.41 & 164 \\
    \textbf{AAPD20} & 52,881 & 45,996 & 4,999 & 944 & 942 & 20 & 1.95 & 165 \\
    \hline
\end{tabular}
}
\end{table}

\subsubsection{Reuters 21578}
The Reuters 215782 corpus\footnote{\url{http://www.daviddlewis.com/resources/testcollections/reuters21578/}} is comprised of documents collected from the Reuters news-wire in 1987.
The documents were manually annotated and the initial version of the data-set was further cleaned up in 1996 resulting in the current form. The training-set contains 7,769 documents,  while 3,019 documents are reserved for testing. For our experiments, to be consistent with the other corpora, we have prepared a calibration and validation sets containing 999 and 1,000 documents  respectively, taken from the training-set. 

\subsubsection{Arxiv Academic Paper Dataset (AAPD)}
AAPD\footnote{\url{https://github.com/lancopku/SGM}}~\cite{yang2018sgm} is a relatively new large data-set for multi-label text classification.
It is composed of abstracts and corresponding subjects of 55,840 papers in the computer science area collected from the arXiv\footnote{\url{https://arxiv.org/}} website.
There are 54 subjects in total and one research paper usually has multiple subjects. This corpus is divided into training, validation and test sets, as described in Table~\ref{tab:corpora_stats}. We have additionally created a calibration set containing 4,999 documents randomly selected from the training part.

\subsubsection{Czech Text Document Corpus v 2.0 (CTDC)}
This corpus\footnote{http://ctdc.kiv.zcu.cz/} consists of real Czech newspaper articles provided by the Czech News Agency~(\v{C}TK)\footnote{https://www.ctk.eu/}. The main part (for training and testing) is composed of 11,955 documents, while the development set contains 2,735 additional articles. The testing protocol assumes a five-fold cross-validation on the main part for training and testing with the use of the development documents as a validation set. The documents belong to different categories such as weather, politics, sport, culture etc. and each document is associated with one or more categories. The corpus was annotated by professional journalists from the Czech News Agency and contains 60 different categories. All documents are automatically morphologically annotated using the UDPipe tool. 

\subsection{Evaluation Metrics}
\label{sec:evaluation_metrics}
The evaluation metrics used for applying CP in multi-label classification can be grouped into two categories. The first one relates to the performance of the forced-prediction mode which enables the comparison of ICP with the underlying model and mostly consists of typical multi-label classification metrics. The second category concerns the set-prediction mode of CP.

In all cases we use the multi-hot representation of $y_i, z_i \in \mathcal{P}(\{\Psi_1,\dots,\Psi_d\})$, where $\vv{y}_i = (t_i^1, \dots, t_i^d) $ corresponds to the true label-set and $\vv{z}_i = (z_i^1, \dots, z_i^d)$ to the prediction label-set for test instance $i\in \{1,\dots,g\}$ which derives from thresholding the raw prediction at $0.5$. Forced-prediction results are evaluated using six different metrics: 

\begin{itemize}
    \item 
    \emph{Classification accuracy $(CA)$} is computed for the complete set of test instances and averaged over their total number. For each instance, a correct prediction is given if and only if the true multi-label target has been fully matched by the highest p-value prediction, i.e. 
    \begin{equation}
        CA = \frac{1}{g}\sum^{g}_{i=1} I(\vv{y}_i=\vv{z}_i),
    \end{equation}{}
    where $I$ is 1 if the condition is true and 0 otherwise. Accuracy is therefore strict and rewards only \emph{absolutely-correct} predictions and not \emph{partially-correct} predictions.

    \item
    The \emph{$F_1$-measure} corresponds to the harmonic mean of precision and recall and its value is in the range [0, 1]. It can be further split into the micro-averaged and macro-averaged types. $F_{micro}$ is averaged over the complete set of test instances, which means that more frequent labels weight more than infrequent ones. Conversely, $F_{macro}$ is first averaged per instance and the results are then averaged over the total number of labels. Consequently, $F_{macro}$ gives equal weights to all labels and it therefore tends to be lower than $F_{micro}$ when poorer performance is observed for the more infrequent ones. The two are defined as:
    \begin{equation}
    \label{eq:fmicro}
        F_{micro} = \frac{2\sum^{d}_{j=1}\sum^{g}_{i=1}{t_i^j} {{z_i^j}}}{\sum^{d}_{j=1}\sum^{g}_{i=1}t_i^j +\sum^{d}_{j=1}\sum^{g}_{i=1}{z_i^j}},
    \end{equation}{}
    \begin{equation}
    \label{eq:fmacro}
        F_{macro} = \frac{1}{d} \sum^d_{j=1} \frac{2\sum^{g}_{i=1}t_i^j z_i^j}{\sum^{g}_{i=1}t_i^j +\sum^{g}_{i=1}z_i^j},
    \end{equation}{}
    \item
    \emph{Hamming Loss (HL)} is evaluated as a loss function and, in contrast to accuracy and $F_1$ measures, the objective is minimisation. It is given by the symmetric difference between actual and predicted labels, averaged over the total number of test instances and it is more suitable than $CA$ for multi-label classification problems as it also rewards \emph{partially-correct} predictions. It is defined as: 
    \begin{equation}
    \label{eq:hl}
        HL = \frac{1}{gd}\sum^g_{i=1}\sum^d_{j=1} t_i^j \oplus z_i^j,
    \end{equation}{} 
    where $\oplus$ is the xor operator.
    
    \item
    We also evaluate \emph{average-confidence ($\overline{Conf.}$)}, which is intended as an overall indication of how likely predictions are compared to all other possible classifications (Section \ref{sec:conformal-prediction}) and it is defined as: 
    \begin{equation}
    \label{eq:conf-criterion}
        \overline{Conf.} = \frac{1}{g}\sum_{i=1}^g 1 - \max_{\vv{Y}_j \neq {\arg\max p_i(\vv{Y}_j)}} p_i(\vv{Y}_j), 
    \end{equation}    
    where we compute the average value of all confidence scores (i.e. $1 -$ the second largest p-value, over all considered label-sets $\vv{Y}_j)$ over $g$ number of test instances.
    \item
    As discussed in Section \ref{sec:conformal-prediction}, credibility indicates how suitable is the training data-set for each test instance. Here we evaluate an overall model suitability using \emph{average-credibility ($\overline{Cred.}$)}, defined as:
    \begin{equation}
    \label{eq:cred-criterion}
        \overline{Cred.} = \frac{1}{g}\sum_{i=1}^g \max_{\vv{Y}_j} p_i(\vv{Y}_j), 
    \end{equation}{}
    where the credibility of example $i$ is the largest p-value out of all considered label-sets $\vv{Y}_j$.
    
\end{itemize}
\bigbreak

The quality of the generated p-values and the practical usefulness of the prediction-sets are evaluated using the following probabilistic criteria\footnote{Probabilistic criteria are conceptually similar to ``proper scoring rules'' in probability forecasting.} identified in \citep{cp-criteria}; for all criteria smaller values are preferable.

\begin{itemize}
    
    \item The \emph{S-criterion (S)}, i.e. Sum criterion, is the average sum of p-values across all test instances and it is independent of significance level $\epsilon$. It is defined as:
    \begin{equation}
    \label{eq:s-criterion}
        S = \frac{1}{g}\sum^{g}_{i=1}\sum_{\vv{Y}_j} p_i(\vv{Y}_j)
    \end{equation}
    
    \item The \emph{OF-criterion (OF)}, i.e. Observed-Fuzziness criterion, is the same as the \emph{S-criterion}, but excluding the p-value of the true label-set. It is defined as:
    \begin{equation}
    \label{eq:of-criterion}
        OF = \frac{1}{g}\sum^{g}_{i=1}\sum_{\vv{Y}_j \neq {\vv{y}_i}} p_i(\vv{Y}_j)
    \end{equation}

    \item The \emph{N-criterion}, i.e. Number criterion, which measures efficiency as a function of significance level. It returns the average number of label-sets across all prediction-sets:
    \begin{equation}
    \label{eq:n-criterion}
        N = \frac{1}{g}\sum^{g}_{i=1}|\Gamma^\epsilon_i|.
    \end{equation}
    where $|\Gamma^\epsilon_i|$ is the size of the resulting prediction-set for instance $i$ at a significance level $\epsilon$. In the  multi-label setting its value ranges between 0 and the number of the possible label-sets $c$.

\end{itemize}

\subsection{Classification Results of Underlying Models}
\label{sec:results_without_icp}
The underlying classifiers are trained on the whole training-set (proper training and calibration sets). We present results on the full data-sets in Table~\ref{tab:results_withput_icp}, as well as on the reduced versions (consisting of only the 20 most frequent labels) in Table~\ref{tab:results_without_icp_20}.

The results indicate that the non-contextualised embeddings based classifiers perform approximately the same while bert surpasses them both, by a large margin. This is in line with results from recent studies that use transformer-based classifiers, like  \cite{stateoftheart20} 
where "MAGNET" is proposed and achieves state-of-the-art performance in multi-label text classification. The authors report F1-micro scores of 0.899 and 0.696 and hamming loss scores of 0.0029 and 0.0252 for the Reuters and AAPD data-sets, respectively. In addition, the state-of-the-art result for the Czech data-set is 0,847, for F1-micro, obtained using the 37 most frequent categories \cite{lenc2016deep}. Based on our experiments and to the best our knowledge, the implementation of bert presented in this study slightly surpasses the state-of-the-art performances obtained by other researchers for the same data-sets.

\begin{table}[ht]
\centering
\footnotesize{
  \caption{Classification results for full data-sets (without ICP).}
  \label{tab:results_withput_icp}
  \begin{tabular}{cc|ccc|ccc|ccc}
    &  & \multicolumn{3}{c|}{\textbf{Reuters}} & \multicolumn{3}{c|}{\textbf{AAPD}} & \multicolumn{3}{c}{\textbf{CTDC}}\\
    & & \textbf{Bert} & \textbf{Randinit} &\textbf{Word2vec} & \textbf{Bert} & \textbf{Randinit} &\textbf{Word2vec} &\textbf{Bert} & \textbf{Randinit} &\textbf{Word2vec} \\
    \hline
    \multicolumn{2}{c|}{\textbf{Accuracy}} & 0.872 & 0.818 & 0.831  & 0.393 & 0.367 & 0.363  & 0.611 & 0.534 & 0.525 \\
    \multicolumn{2}{c|}{\textbf{F1-micro}} & 0.907 & 0.854 & 0.866  & 0.735 & 0.704 & 0.700  & 0.864 & 0.825 & 0.824 \\
    \multicolumn{2}{c|}{\textbf{F1-macro}} & 0.557 & 0.429 & 0.436  & 0.592 & 0.506 & 0.471  & 0.806 & 0.712 & 0.703 \\
    \multicolumn{2}{c|}{\textbf{Hamming loss}} & 0.003  & 0.004 & 0.004  & 0.023 & 0.024 & 0.024 & 0.012 & 0.015 & 0.015 \\
    \hline
\end{tabular}
}
\end{table}

\begin{table}[ht]
\centering
\footnotesize{
  \caption{Classification results for reduced data-sets (without ICP).}
  \label{tab:results_without_icp_20}
  \begin{tabular}{cc|ccc|ccc|ccc}
    &  & \multicolumn{3}{c|}{\textbf{Reuters20}} & \multicolumn{3}{c|}{\textbf{AAPD20}} & \multicolumn{3}{c}{\textbf{CTDC20}}\\
    & & \textbf{Bert} & \textbf{Randinit} &\textbf{Word2vec} & \textbf{Bert} & \textbf{Randinit} &\textbf{Word2vec} &\textbf{Bert} & \textbf{Randinit} &\textbf{Word2vec} \\
    \hline
    \multicolumn{2}{c|}{\textbf{Accuracy}} & 0.911 & 0.882 & 0.883  & 0.565 & 0.539 & 0.531  & 0.703 & 0.647 & 0.636 \\
    \multicolumn{2}{c|}{\textbf{F1-micro}} & 0.942 & 0.923 & 0.924  & 0.804 & 0.766 & 0.767  & 0.874 & 0.850 & 0.842 \\
    \multicolumn{2}{c|}{\textbf{F1-macro}} & 0.879 & 0.839 & 0.841  & 0.745 & 0.692 & 0.699  & 0.877 & 0.847 & 0.839 \\
    \multicolumn{2}{c|}{\textbf{Hamming loss}} & 0.007 & 0.009 & 0.009  & 0.037 & 0.042 & 0.042 & 0.025 & 0.029 & 0.030 \\
    \hline
\end{tabular}
}
\end{table}




For all experiments apart from CTDC20, F1-macro is much lower than F1-micro which suggests that the classification models have poorer performance on the less frequent classes. For the bert classifier, the difference between F1-micro and F1-macro is less than that of the non-contextualised classifiers, meaning that bert is more efficient in classifying rarely encountered labels. Hamming loss, on the reduced data-sets, shows a small advantage of bert over non-contextualised classifiers in predicting fewer false positive labels. For all classifiers, best performance (accuracy and F1-micro) is achieved on the data-sets of Reuters and Reuters20, which is significantly better than the other data-sets and it is a relatively good performance in general for a multi-label classification tasks. This can be, at least partly, attributed to the average number of labels per instance (Reuters20: $1.16$ and Reuters: $1.24$) making its complexity very close to that of a multi-class problem.

For the CNN classifiers, the initialisation of the embedding layer with word2vec embeddings does not seem to add any performance gains in comparison to the random initialisation when the embedding layer and the task-specific layers are trained (fine-tuned for the case of word2vec) simultaneously. On the contrary, the word2vec classifier performs slightly worst than randinit. This is consistent with the results from \cite{lenc2017word}, where authors concluded that for longer texts better results are achieved using embeddings initialised randomly and trained jointly with the classification network.

Classification results for the reduced data-sets are significantly better compared to the results for the full versions. This is as expected based on the fact that the reduced data-sets actually correspond to an easier classification problem (fewer number of possible labels and label-combinations). For the AAPD20 and CTDC20 data-sets, that have very similar average number of labels per instance ($1.96$ and $1.95$, respectively), there is a significant difference in accuracy and F1-micro in favour of CTDC20, despite the fact that AAPD20's total number of training instances is almost 5 times greater. A possible explanation for this -- apart from the distribution of labels -- is that in the case of CTDC20 the average number of words per document is much larger (269 over 165), thus providing richer feature inputs to the model.\par

\subsection{Original LP-ICP}
\label{sec:original_lp_icp}
We apply LP-ICP based on the three classifiers on the data-sets of Reuters20, AAPD20 and CTDC20. This section presents the results obtained from the forced-prediction mode, used here for assessing the impact of ICP on the classification performance, and from the set-prediction mode that realises the confidence information benefits. 

\subsubsection{Forced Prediction Results}
We present forced-prediction results in Table \ref{tab:forced_prediction_results_with_lp_icp_on_reduced_data-sets}, estimated from the label-sets with the highest p-value, as described in Section \ref{sec:conformal-prediction}. In combination with Table~\ref{tab:results_without_icp_20}, a direct performance comparison between the underlying classifiers with and without the use of ICP is possible. This shows that the performance of forced-prediction is negligibly different from that of the non-ICP classification for all data-sets, and indicates that no substantial classification performance is sacrificed by the use of ICP. Also, the confidence information provided from $Conf.$, which serves as an indicator of the likelihood that the predicted label-set is the true target,  is, on average high. \par

\begin{table}[ht]
\centering
\footnotesize{
  \caption{Forced prediction results with LP ICP on reduced Data-sets}
  \label{tab:forced_prediction_results_with_lp_icp_on_reduced_data-sets}
  \begin{tabular}{cc|ccc|ccc|ccc}
    &  & \multicolumn{3}{c|}{\textbf{Reuters20}} & \multicolumn{3}{c|}{\textbf{AAPD20}} & \multicolumn{3}{c}{\textbf{CTDC20}}\\
    & & \textbf{Bert} & \textbf{Randinit} &\textbf{Word2vec} & \textbf{Bert} & \textbf{Randinit} &\textbf{Word2vec} &\textbf{Bert} & \textbf{Randinit} &\textbf{Word2vec} \\
    \hline
    \multicolumn{2}{c|}{\textbf{Accuracy}} & 0.913 & 0.882 & 0.881  & 0.553 & 0.533 & 0.534  & 0.691 & 0.636 & 0.635 \\
    \multicolumn{2}{c|}{\textbf{F1-micro}} & 0.944 & 0.920 & 0.920  & 0.792 & 0.767 & 0.774  & 0.872 & 0.842 & 0.841 \\
    \multicolumn{2}{c|}{\textbf{F1-macro}} & 0.885 & 0.813 & 0.826  & 0.724 & 0.691 & 0.705  & 0.873 & 0.840 & 0.838 \\
    \multicolumn{2}{c|}{\textbf{Hamming loss}} & 0.006  & 0.009 & 0.009  & 0.040 & 0.042 & 0.043 & 0.025 & 0.030 & 0.030 \\
    
    \hline
    \multirow{3}{*}{$\overline{Conf.}$}
    &\textbf{$L_{2}$} & 0.956 & 0.936 & 0.928  & 0.658 & 0.623 & 0.615  & 0.787 & 0.739  & 0.730 \\
    &\textbf{$L_{4}$} & 0.967 & 0.941 & 0.946  & 0.701 & 0.661 & 0.665  & 0.806 & 0.758  & 0.751 \\
    &\textbf{$L_{8}$} & 0.970 & 0.951 & 0.950  & 0.721 & 0.687 & 0.671  & 0.816 & 0.771  & 0.767 \\

    \hline
    \multirow{3}{*}{$\overline{Cred.}$}
    &\textbf{$L_{2}$} & 0.533 & 0.542 & 0.550  & 0.644 & 0.655 & 0.669  & 0.577 & 0.597 & 0.602 \\
    &\textbf{$L_{4}$} & 0.529 & 0.543 & 0.550  & 0.650 & 0.671 & 0.673  & 0.577 & 0.598 & 0.603 \\
    &\textbf{$L_{8}$} & 0.527 & 0.601 & 0.607  & 0.651 & 0.663 & 0.674  & 0.577 & 0.602 & 0.605 \\

    \hline
\end{tabular}
}
\end{table}

In all cases $\overline{Conf.}$, is relatively high considering the huge number of possible label-sets, 
which is also reflected in the corresponding accuracy values. The very high values of both accuracy and $\overline{Conf.}$ for the Reuters20 data-set are attributed to its lower complexity due to the low cardinality of its label-sets. $\overline{Cred.}$ is sufficiently high, indicating that the trained models are suitable for classifying the test instances. 

\subsubsection{Prediction Sets Results}
\label{sec:prediction_sets_results_reduced_data_sets}

Lower values are preferred for the $S$ and $OF$ criteria as this indicates that the average p-values of $\vv{Y}_j$ among all instances are kept small, leading to narrower prediction-sets (smaller $N$ criterion). The results for the $S$ and $OF$ criteria are presented in Table \ref{tab:s_and_of_criteria_lp_icp}. In all cases, the $S - OF$ difference is around 0.5 as expected, since the average p-values of the true labels are expected to be 0.5. The lower $S$ criterion values are obtained from $L_8$ indicating that a high p factor for $L_p$ norm increases the ICP performance.

\begin{table}[ht]
\centering
\footnotesize{
    \caption{S $\&$ OF criteria for ICP in reduced data-sets}
    \label{tab:s_and_of_criteria_lp_icp}
    \begin{tabular}{cc|ccc|ccc|ccc}
    &  & \multicolumn{3}{c|}{\textbf{Reuters20}} & \multicolumn{3}{c|}{\textbf{AAPD20}} & \multicolumn{3}{c}{\textbf{CTDC20}}\\
    & Criterion & \textbf{Bert} & \textbf{Randinit} &\textbf{Word2vec} & \textbf{Bert} & \textbf{Randinit} &\textbf{Word2vec} &\textbf{Bert} & \textbf{Randinit} &\textbf{Word2vec} \\
    
    \hline
    \multirow{2}{*}{$L_{2}$}
    &\textbf{$S$} & 144.180 & 148.100 & 150.350 & 71.199 & 60.760 & 105.653  & 182.105 & 171.714  & 181.102 \\
    &\textbf{$OF$} & 143.655 & 147.573 & 149.817 & 70.697 & 60.250 & 105.144  & 181.606 & 171.220 & 181.102 \\


    
    \hline
    \multirow{2}{*}{$L_{4}$}
    &\textbf{$S$} & 142.828 & 148.830 & 146.810 & 56.068 & 48.366 & 90.925  & 169.456 & 173.001 	& 161.488 \\
    &\textbf{$OF$} & 142.308 & 148.304 & 146.276 & 55.567 & 47.857 & 90.416  & 168.957 &	 172.502 & 160.987 \\



    \hline
    \multirow{2}{*}{$L_{8}$}
    &\textbf{$S$} & 142.803 & 150.012 & 147.337 & 50.807 & 41.974 & 90.860  & 158.305 &	166.664 	& 153.257 \\
    &\textbf{$OF$} & 142.285 & 149.434 & 146.746 & 50.306 & 41.465 & 90.353  & 157.806	& 166.166 &	152.754  \\



    
    \hline
\end{tabular}
}
\end{table}

\subsubsection*{N Criterion}
\begin{figure}[!hp]
      \centering
      \includegraphics[width=\textwidth,height=\textheight,keepaspectratio]{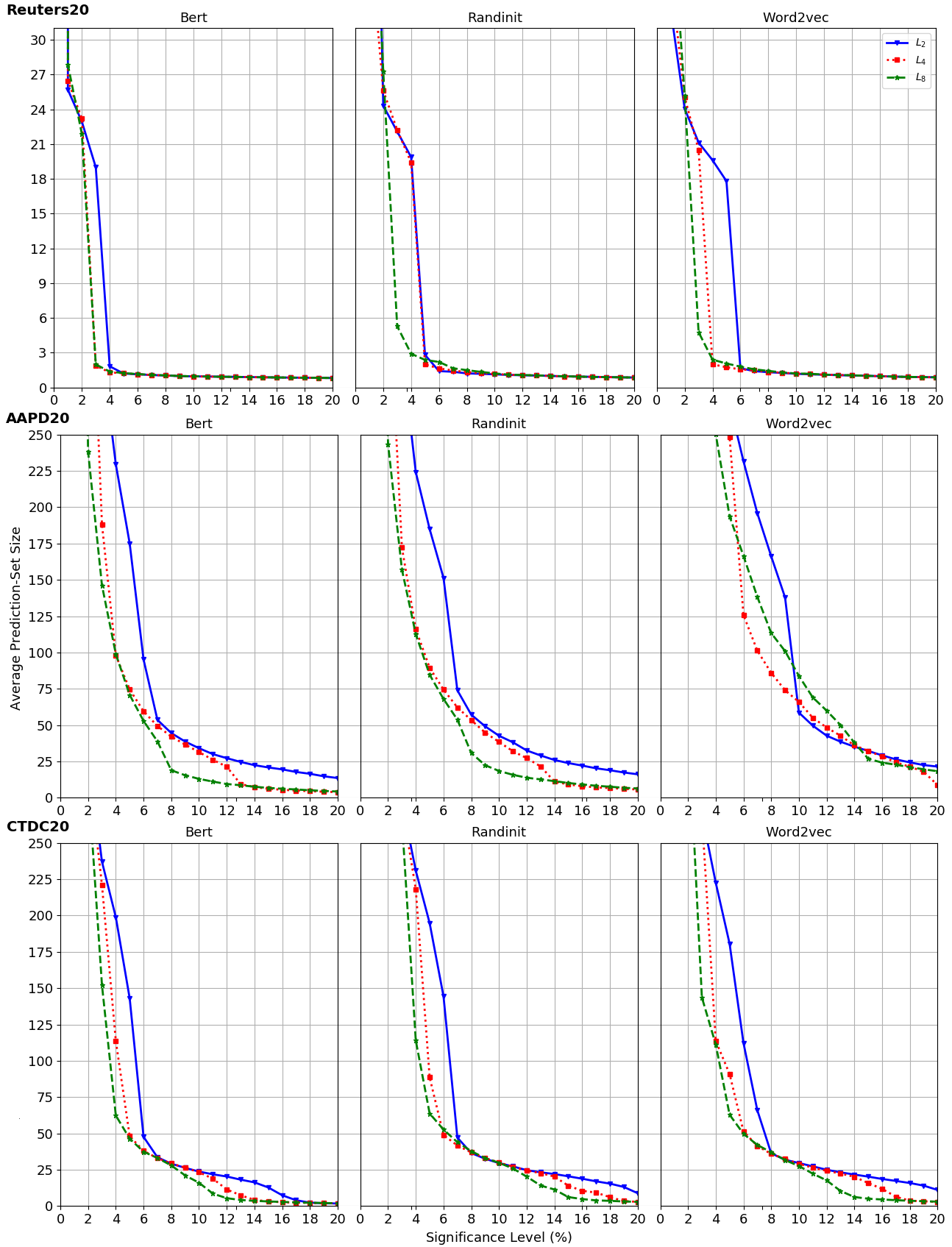}
      \caption{N criterion (y-axis) for Reuters20, AAPD20 and CTDC20 data-sets per significance level (x-axis).}
      \label{fig:n_reduced_datasets}
      \vspace{-15pt}
    \end{figure}

The values of the N-criterion for Reuters20, AAPD20 and CTDC20 data-sets are presented in Figure \ref{fig:n_reduced_datasets}. Given that the error rate is guaranteed within the CP framework, low N-criterion values are preferred as they indicate practically useful prediction sets. For all cases examined, nonconformity metric $L_2$ marks the worst performance, while the best performance is mostly achieved with $L_8$. Based on this observation we can conclude that the increase of the p factor in $L_p$ norm leads to tighter prediction sets, also supported from  the observations pointed out for the $S$ and $OF$ criteria. \par

On Reuters20, the best performance is achieved using the nonconformity metrics $L_4$ and $L_8$. For significance level, $\epsilon = 0.01$, classifiers bert, randinit and word2vec produce, on average, prediction sets of 13, 36 and 20 label-sets respectively. For $\epsilon = 0.05$, the corresponding values are 1.22 for bert, 1.85 for randinit and 1.67 for word2vec. The prediction sets are tight enough to be useful in practise for high confidence levels ($\geq 0.95)$, especially when considering the total number of possible combinations, $c(20,7) = 137979$.\par

On the AAPD20 and CTDC20 data-sets the best performances achieved, for almost all cases, were with nonconformity metric $L_8$. For $\epsilon = 0.05$ on the AAPD20 data-set, classifiers bert, randinit and word2vec marked $N$ criterion values of 50, 118 and 178, respectively, while the corresponding values for the CTDC20 data-set were 46, 64 and 63. Although the total number of possible combinations is comparable with Reuters20 (for AAPD20 is  $c(20,6) = 60459$ and for CTDC20 is $c(20,7) = 137979$) the prediction set sizes are significantly larger. This can be attributed to the poorer performance of the underlying classifiers on the corresponding data-sets.\par

Prediction sets obtained from the bert classifier were significantly tighter compared to randinit and word2vec, for significance levels $\leq 0.05$. Additionally, prediction set sizes of randinit and word2vec were approximately the same. We can therefore conclude that prediction sets of ``better'' classifiers, in terms of performance, are tighter, as was implied from the ICP framework. 

We also calculate the median prediction set sizes (not shown for space consideration). In almost all cases, results show that these numbers are smaller than or equal to the N criterion, for significance levels $\leq 0.05$, which points to extreme values of prediction set sizes for specific test instances and high confidence levels, especially for the AAPD20 data-set.

\subsubsection*{Empirical Error Rate}

\begin{figure}[!p]

  \centering
  \includegraphics[width=\textwidth,height=\textheight,keepaspectratio]{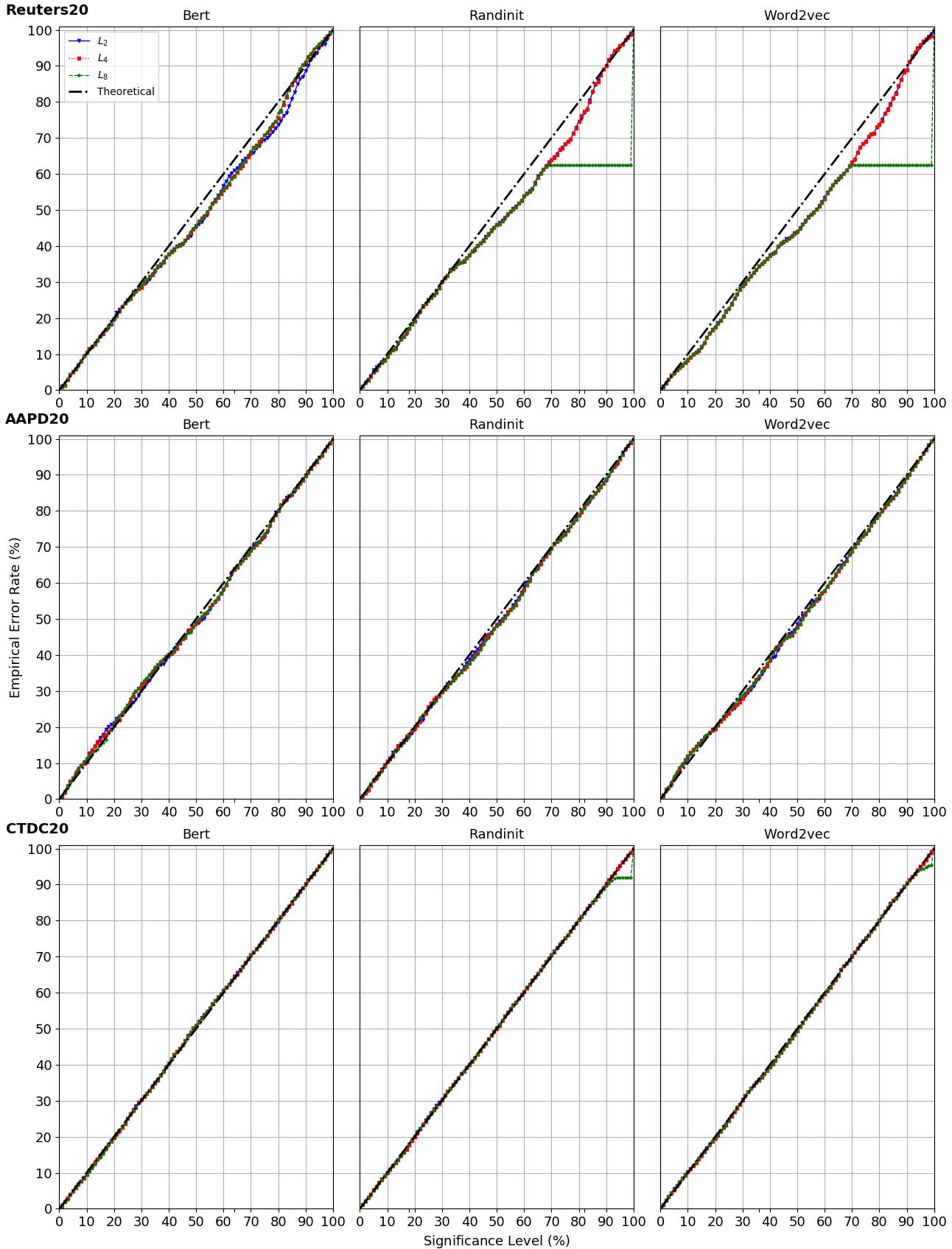}
  \vspace{-15pt}
  \caption{Empirical Error Rate (y-axis) for Reuters20, AAPD20 and CTDC20 data-sets per significance level (x-axis).}
  \label{fig:errors_reduced_data_sets}
\end{figure}

We also examine the empirical validity of our conformal predictors, by plotting the test-set error-rate against  the  significance  level  in  the  range  $[0, 1]$. Results for the data-sets: Reuters20, AAPD20 and CTDC20 are presented in
Figure~\ref{fig:errors_reduced_data_sets}. It is shown that, as  guaranteed theoretically, the error-rate is always less than or equal to the significance level (up to statistical fluctuations) and confirms that our data satisfies the exchangeability assumption. More specifically, for AAPD20 and CTDC20 datasets, the empirical error rate, especially for low significance levels, is close to the theoretical as expected from the LP implementation of CP. 

Results reveal an interesting observation for nonconformity metric $L_8$ and significance levels $\geq 0.7$. There are cases in the Reuters20 and CTDC20  data-sets where the empirical error-rate deviates from the diagonal (but still valid). A possible explanation is that the $L_8$ norm tends to act like $L_{\inf}$ norm, i.e. it only takes into account the label with the larger prediction error neglecting information from other labels. Therefore, there is a dense concentration of nonconformity scores $\alpha_i$ of the calibration instances $x_i$ around specific values, causing the same dense concentration of p-values of $\vv{Y}_j$ as well.

\subsection{Efficient LP-ICP}
\label{sec:efficient_icp_power_set_in_full_data_set}
To the best of our knowledge this study is the first one that presents experimental results of a CP approach for such demanding multi-label classification problems. In this section, we first discuss the computational complexity improvements of the efficient LP-ICP compared to the original approach. Next, we evaluate its performance in the forced-prediction mode and finally we evaluate the resulting set-predictions, both in terms of tightness and empirical error rate.

\subsubsection{Computational Complexity}
\label{sec:Computational_complexity}
\begin{figure}[ht]
  \centering
  \includegraphics[width=0.9\textwidth,height=0.9\textheight,keepaspectratio]{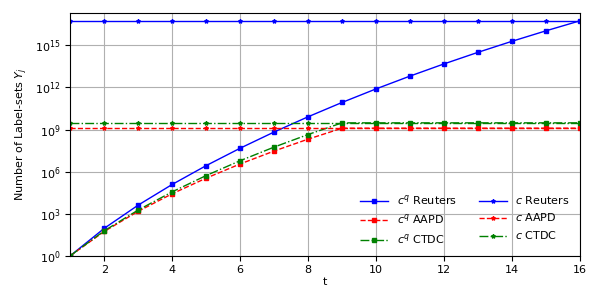}
  \caption{Theoretical number of possible label-sets $Y_j$ considered for a single instance in case of the original LP-ICP ($c$) and of the proposed efficient LP-ICP ($c^q$). Y-axis is in log scale.}
  \label{fig:computation_load_efficient_vs_original}
\end{figure}

The computational load for a single instance, depicted in Figure \ref{fig:computation_load_efficient_vs_original}, is expressed as the number of label-sets considered and evaluated, which is $c = c(d,l)$ and $c^q = |Q(t_{n+m})|, Q(t_{n+m}) = \{Y_j: |Y_j  \triangle z_{n+m}| < t_{n+m}\}$ for the original and the efficient LP-ICP respectively (see Section \ref{sec:efficient_lp_icp}). In the case of the original LP-ICP $c$ is dataset depended (because of the $d$ and $l$) therefore $c$ is constant across all instances, while $c^q$ is additionally dependent on the $t = t_{n+m}(\alpha_{i_0}(\epsilon))$ of each instance $x_{n+m}$. For $t \leq l$, $c > c^q$ and $c = c^q$ for $t > l$. As shown in Figure \ref{fig:computation_load_efficient_vs_original}, the computational complexity of the proposed efficient LP-ICP for a single instance decreases exponentially as $t$ decreases linearly. The number of unique labels $d$ and the maximum observed cardinality $l$ of the  data-set are: $d=90$ and $l =15$ for Reuters; $d=54$ and $l = 8$ for AAPD; and $d=60$ and $l =8$ for CTDC.

The total computation load for the original ($c_{total}$) and the proposed ($c^q_{total}$) LP-ICP approaches is presented in Table~\ref{tab:computational_demand}. As $c$ is the same for all test instances, $c_{total} = c*g$ with $g$ being the number of instances while $c^q_{total} = \sum^{g}_{i=1} |Q(t_{n+i})|$ i.e. the total number of label-sets considered across all test instances. Results show that the number of labels-sets evaluated by the efficient LP-ICP are orders of magnitude less than those evaluated by the original LP-ICP. As an example of the computational performance gains, consider bert on the Reuters data-set with the L2 metric. In this case, the total number of $g=3019$ test instances requires considering  $2e+6$ label-sets, while the corresponding original LP-ICP requires $5.68e+16 * 3019  \approx 1.7e+19$. 

\begin{table}[t!]
\centering
\footnotesize{
    \caption{Computational demand for Reuters, AAPD and CTDC data-sets for significance level $\epsilon = 5\%$.  Total $c$ is the sum of all labels-sets considered across all test instances for the LP-ICP while $c^q$ is the corresponding one for efficient LP-ICP.} 
    \label{tab:computational_demand}
  \begin{tabular}{cL|ccc|ccc|ccc}
    &  & \multicolumn{3}{c|}{\textbf{Reuters}} & \multicolumn{3}{c|}{\textbf{AAPD}} & \multicolumn{3}{c}{\textbf{CTDC}}\\
    \hline
    & & \textbf{Bert} & \textbf{Randinit} &\textbf{Word2vec} & \textbf{Bert} & \textbf{Randinit} &\textbf{Word2vec} &\textbf{Bert} & \textbf{Randinit} &\textbf{Word2vec} \\
    \hline
    
    \hline
    \multirow{4}{*}{$L_2$}
    &\textbf{Average t} & 2.09 & 2.13 & 2.18 & 4.49 & 4.58 & 4.61 &	3.66 & 4.01 & 4.05 \\
    &\textbf{Median t} &2 & 2 & 2 & 4 & 4 & 4 & 4 & 4 & 4 \\
    &\textbf{$c^q_{total}$} & 2e+6 & 4.4e+6 &	9.1e+6 &  1.1e+9 & 7.6e+8& 7.5e+8 & 6.7e+8 &	4e+8 & 4.6e+8\\
    &\textbf{Instances with Prohibitive $t$  ($\%$)} & 0 & 0 & 0 & 0 & 0 & 0 & 0 & 0 & 0 \\

    \hline
    \multirow{4}{*}{$L_4$}
    &\textbf{Average t} & 1.59 & 2.13 & 2.19 & 4.53 & 4.60 & 4.50 & 3.62 & 3.92 & 3.90 \\
    &\textbf{Median t} & 1 & 2 & 2 & 5 & 4 & 4 & 3 & 4 & 4 \\
    &\textbf{$c^q_{total}$} & 3.7e+8 & 6.1e+7 & 9.2e+7 & 3.7e+9 &	3.6e+9 &	2.6e+9 & 3.4e+9 & 3.3e+9 & 3.3e+9\\
    &\textbf{Instances with Prohibitive $t$ ($\%$)} & 0 & 0 & 0 & 0.6 & 0.8 & 0.4 & 0.13 & 0.04 & 0.04 \\
    
    \hline
    \multirow{4}{*}{$L_8$}
    &\textbf{Average t} & 1.57 & 2.18 & 2.19 & 4.65 & 5.13 & 4.78 & 3.77 &	4.00 & 3.98 \\
    &\textbf{Median t} & 1 & 2 & 2 & 4 & 5 & 5 & 3 & 4 & 4 \\
    &\textbf{$c^q_{total}$} & 6.2e+10 & 9.3e+8 &	1.8e+9 & 1.2e+10 & 1e+10 & 1.2e+9 &2.3e+10 & 2.2e+10 & 2.2e+10 \\
    &\textbf{Instances with Prohibitive $t$  ($\%$)} & 0.50 & 	0.03  & 0.07 & 	3.6 & 3.3 & 2.3 & 0.40  & 0.53 & 0.66\\
    \hline
    
    &\textbf{$c_{total}$} & &  1.7e+19 & & & 1.2e+12& & & 7.1e+12 & \\
    \hline

\end{tabular}
}
\end{table}

For all cases, average and median $t$, also shown in Table~ \ref{tab:computational_demand}, is kept relatively small resulting in significantly lower computation burden for most instances than the one in original LP-ICP. For a typical instance on Reuters dataset $t \leq 3$ and the average computational load is $c^q \leq |Q_{Reuters}(3)| = 4095$ (see Figure \ref{fig:computation_load_efficient_vs_original}) while $c_{Reuters} \approx 5.68e+16$. According to the average $t$, for CTDC $c^q \leq |Q_{CTDC}(5)| \approx 5.2e+5$ while $c_{CTDC} \approx 3e+9$; for AAPD $c^q \leq |Q_{AAPD}(5)| \approx 3.4e+5$ while $c_{AAPD} \approx 1.2e+9$. Our implementation of efficient LP-ICP on python 3.6 enables us to run the process with an average speed of $13e+4$ label-sets/second per cpu core. This leads, to a few seconds of computational time per cpu core for most test instances. 

Despite the efficiency improvement of the proposed LP-ICP approach, there are still some instances with high $t$ for which the computational load is too high for our hardware capabilities because of RAM limitations. We were nevertheless able to obtain results for up to $c^q \approx 5.6e+7$, which corresponds to $t \leq 6$ for Reuters and $t \leq 7$ for AAPD and CTDC data-sets. The percentage of test instances with $c^q > 5.6e+7$, shown in Table \ref{tab:computational_demand} as ``Instances with Prohibitive $t$ (\%)'', is significantly low, less than $1\%$ for almost all cases (expect $L_8$ on AAPD).

\subsubsection{Forced Prediction Results}
\label{sec:efficient-forced-prediction-results}
Table~\ref{tab:forced_prediction_results_efficient} summaries the forced predictions results, i.e. label-sets with the highest p-values. Comparing these values with those reported in Table~\ref{tab:results_withput_icp} 
we conclude that there is an insignificant performance loss from the use of ICP. The $Conf.$ values provided by ICP are on average relatively high considering the corresponding accuracy. Additionally, the underlying models are shown to be suitable for the test instances , as $\overline{Cred.}$ is sufficiently large. 

\begin{table}[ht]
\centering
\footnotesize{
  \caption{Results with efficient ICP power-set - forced prediction}
  \label{tab:forced_prediction_results_efficient}
  \begin{tabular}{cc|ccc|ccc|ccc}
    &  & \multicolumn{3}{c|}{\textbf{Reuters}} & \multicolumn{3}{c|}{\textbf{AAPD}} & \multicolumn{3}{c}{\textbf{CTDC}}\\
    & & \textbf{Bert} & \textbf{Randinit} &\textbf{Word2vec} & \textbf{Bert} & \textbf{Randinit} &\textbf{Word2vec} &\textbf{Bert} & \textbf{Randinit} &\textbf{Word2vec} \\
    \hline
    \multicolumn{2}{c|}{\textbf{Accuracy}} & 0.866 & 0.833 & 0.819  & 0.403 & 0.361 & 0.359  & 0.593 & 0.516 & 0.514 \\
    \multicolumn{2}{c|}{\textbf{F1-micro}} & 0.899 & 0.866 & 0.856  & 0.735 & 0.693 & 0.689  & 0.856 & 0.816 & 0.815 \\
    \multicolumn{2}{c|}{\textbf{F1-macro}} & 0.489 & 0.459 & 0.420  & 0.566 & 0.492 & 0.499  & 0.792 & 0.696 & 0.690 \\
    \multicolumn{2}{c|}{\textbf{Hamming loss}} & 0.004  & 0.004 & 0.004  & 0.022 & 0.025 & 0.026 & 0.013 & 0.016 & 0.016 \\
    
    \hline
    \multirow{2}{*}{$\overline{Conf.}$}
    &\textbf{$L_{2}$} & 0.923 & 0.895 & 0.887  & 0.503 & 0.449 & 0.438  & 0.676 & 0.589  & 0.594 \\
    &\textbf{$L_{4}$} & 0.942 & 0.905 & 0.898  & - & - & -  & - & -  & - \\

    \hline
    \multirow{2}{*}{$\overline{Cred.}$}
    &\textbf{$L_{2}$} & 0.548 & 0.548 & 0.553  & 0.694 & 0.737 & 0.748  & 0.627 & 0.656 & 0.656 \\
    &\textbf{$L_{4}$} & 0.548 & 0.548 & 0.560  & - & - & -  & - & - & - \\

    \hline
\end{tabular}
}
\end{table}

\subsubsection{Prediction-set Results}
\label{sec:efficient-prediction-set-results}
In this section we present the N criterion and the empirical error rate results\footnote{For the efficient LP-ICP version the S and Of criteria can not be evaluated as p-values are only calculated for the promising label-sets.}  for the computationally affordable cases of  nonconformity measures $L_2$ for all data-sets and $L_4$ for Reuters. 

\subsubsection*{N criterion}
\label{sec:n_criterion}

\begin{figure}[p]
  \centering
  \includegraphics[width=\textwidth,keepaspectratio]{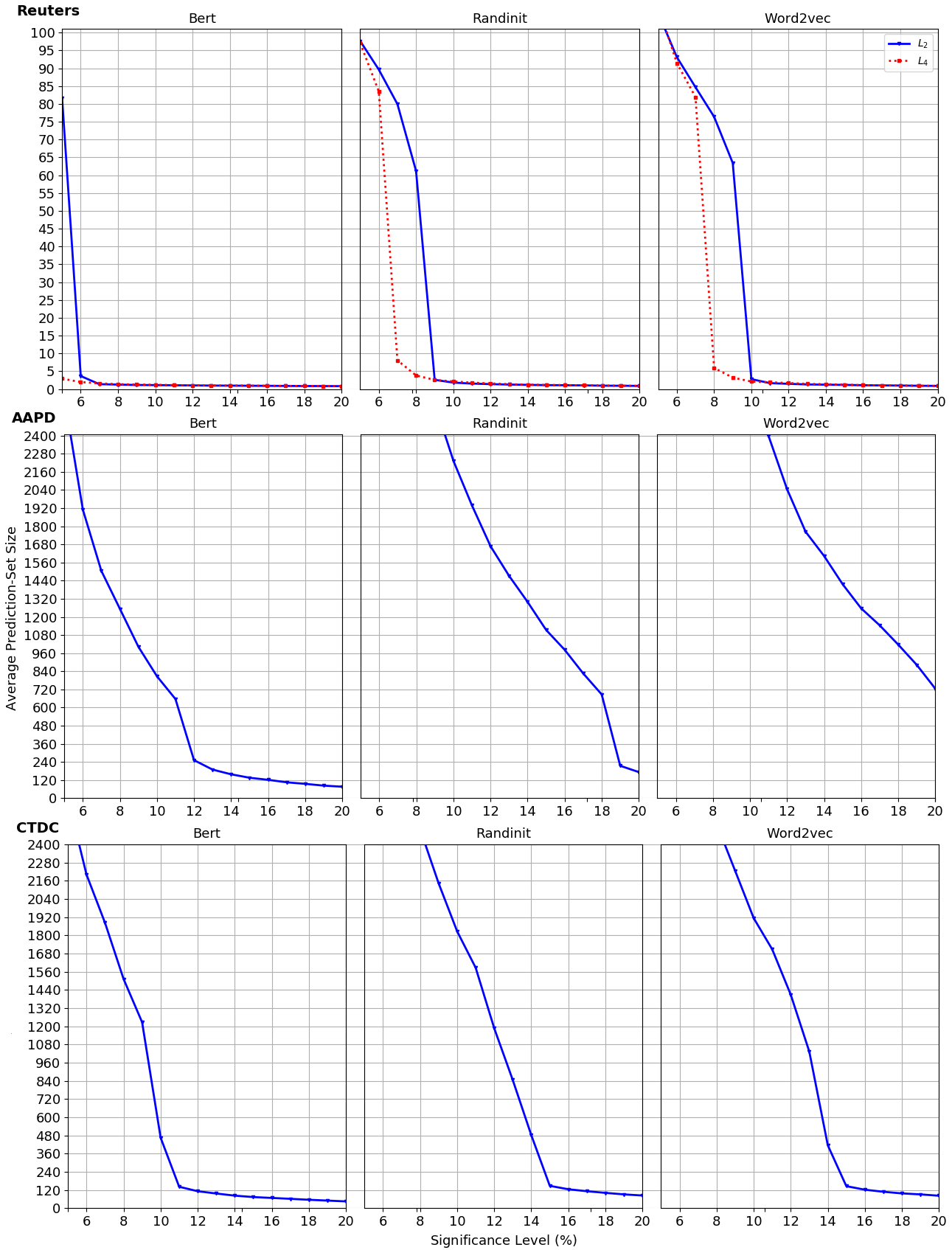}
  \vspace{-15pt}
  \caption{N criterion (y-axis) for Reuters, AAPD and CTDC datasets per significance level (x-axis)}
  \label{fig:n_full_datasets}
\end{figure}

The N criterion results for Reuters, AAPD and CTDC data-sets for confidence level up to $0.95$ are depicted in Figure~\ref{fig:n_full_datasets}. As expected, the N criterion performance  depends greatly on the performance of the underlying classifiers and the complexity of the multi-label classification problem. Thus, best results were obtained with the bert classifier on Reuters. In all cases, the N criterion values for bert are lower than the ones for randinit and word2vec, for the same significance levels. 
In line with the experiment results on reduced data-sets in Section \ref{sec:prediction_sets_results_reduced_data_sets}, higher values of factor $p$ lead to narrower prediction sets.

On Reuters for $\epsilon = 0.05$ we obtained average prediction set size of $2.93$ which is tight enough to be useful in practise, especially considering that the number of possible label-sets is $c\approx 5.68e+16$. The convergence of the CNN classifiers for $L_2$ to an N criterion level of around 3 label-sets is achieved for $\epsilon \approx 0.1$. 

On the AAPD and CTDC data-sets, which are more difficult classification problems than Reuters, the average prediction set size is arguably a good result given the low accuracy of the state-of-the-art underlying models. For example, in the case of AAPD with bert we obtain a prediction set with 76 label-sets out of the possible 1.2e+9 for a confidence of $80\%$ which is double the accuracy of the underlying model ($40\%$).

\begin{figure}[p]
  \centering
  \includegraphics[width=\textwidth,height=\textheight,keepaspectratio]{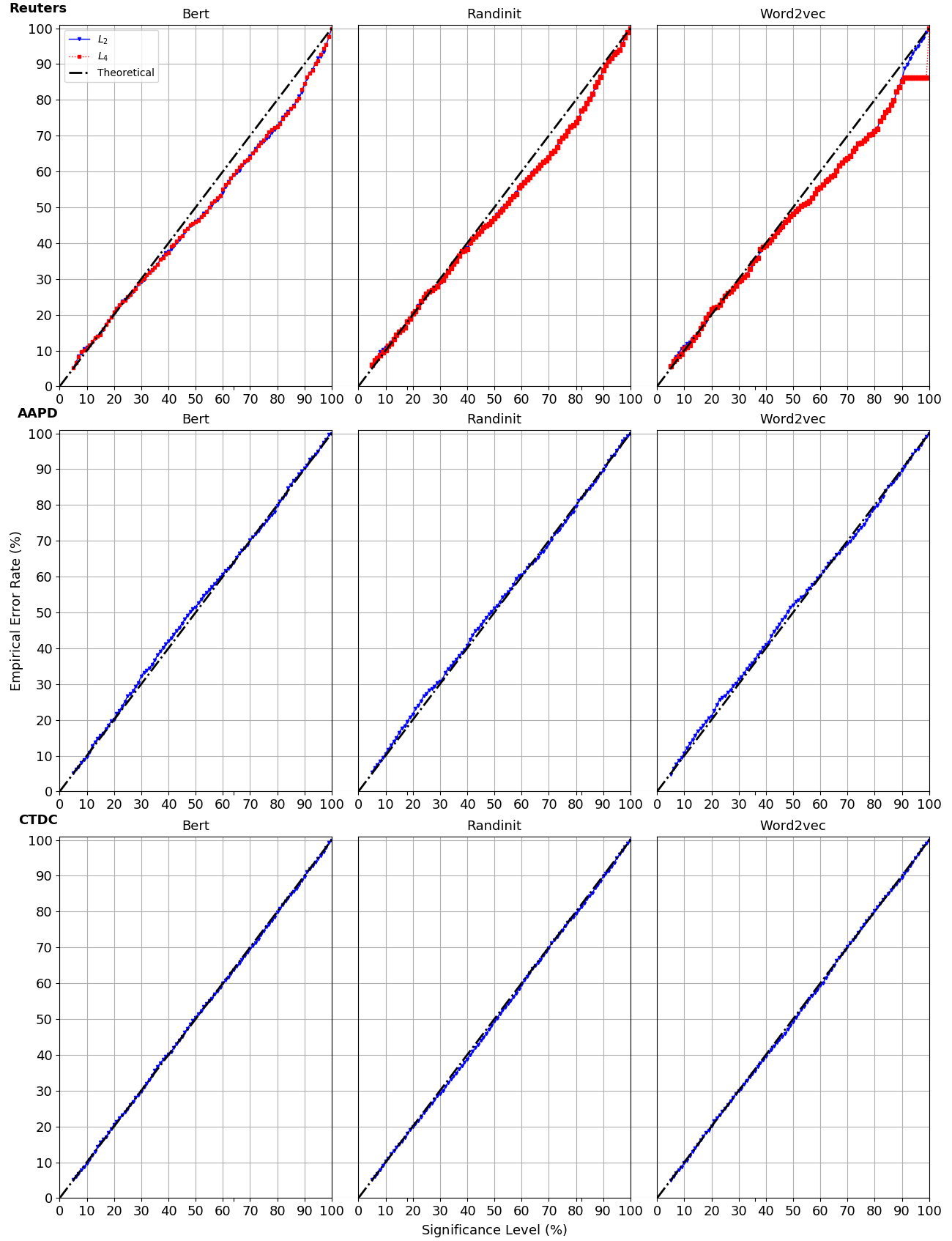}
  \vspace{-11pt}
  \caption{Empirical Error Rate (y-axis) for Reuters, AAPD and CTDC datasets per signifiance level (x-axis).}
  \label{fig:errors_full_datasets}
\end{figure}

\subsubsection*{Empirical Error Rate}
The test error-rate of the prediction sets obtained per significance level $[0.05,1]$ for all data-sets is presented in Figure \ref{fig:errors_full_datasets}. The error-rate across all data-sets is less than or equal to the significance level up to statistical fluctuations, as theoretically guaranteed. The error-rate in the case of AAPD is more prone to fluctuations that slightly surpass the significant level. An explanation for this is the complexity of the specific multi-label classification problem. As possible classifications $Y_j$ grow exponentially when the number of unique labels increases, the calibration set may get less suitable as a reference set for the test set.

\section{Conclusions and Future Work}

In this study we examine the application of CP on the problem of multi-label text classification and assess the performance of different approaches on subsets of two English-language and one Czech-language data-sets. Specifically, we apply the inductive version of CP using the Label Powerset problem formulation under three different versions of a nonconformity measure. We experiment with three underlying classifiers: (i) bert; (ii) randinit; and (iii) word2vec, investigating the performance differences between contextualised (in (i)) and non-contextualised (in (ii) \& (iii)) WEs. Additionally, we address the computationally demanding nature of the task by proposing a variation of the LP-ICP implementation that eliminates from consideration a significant number of label-sets that would surely not be included in the prediction sets. We prove that the efficient LP-ICP approach fully respects the CP guarantees.

Our results indicate that the bert classifier is considerably better performing compared to classifiers using non-contextualised embeddings, for the task of multi-label text classification. To the best of our knowledge our bert implementation slightly overcame the-state-of-the-art performance for all data-sets employed in this study. 

The good performance of our classifiers is carried-on to the forced prediction mode of ICP without any substantial loss in classification accuracy. This shows that the removal of the calibration instances required by ICP does not cause a significant impact on performance.

In the set-prediction mode the best results were achieved for the Reuters20 and Reuters data-sets. The prediction sets obtained are tight enough to be practically useful even at high confidence levels ($\geq$ 0.95). Even though the prediction sets for equally high confidence levels are not as tight for the rest of the data-sets (AAPD20, AAPD, CTDC20 and CTDC), usable prediction sets are obtained at lower confidence levels, which are still much higher than the state-of-the-art accuracy on these data-sets. Regarding nonconformity measures, the use of large factor $p > 2$ for $L_p$ norms produces tighter prediction sets, but increases the computational burden of the efficient LP-ICP. 

Using our efficient LP-ICP implementation we were able to apply CP on data-sets with high numbers of labels, which were previously infeasible to process, at least with moderately powerful hardware set-ups. Our results show that this approach can be applied to equally challenging data-sets and provide practically useful prediction sets. Furthermore, although we specifically apply the efficient version of LP-ICP to text classification, it is worth mentioning that it can be utilised in any multi-label classification scenario.

We intent to expand our work on the efficient ICP prediction set computation with the aim of further improving efficiency. Furthermore, as our current study is limited to the $L_p$ norms, we intent to extend it to more sophisticated nonconformity measures that also take into account label correlation, such as the one used in our previous work \cite{paisios2019deep}. This will allow us to benefit in full from the Label Powerset approach. Finally, we plan to investigate the refinement of the prediction set outputs in order to make them easier to interpret.

\section*{Acknowledgements}

     We thank the editor and the anonymous reviewers for their  insightful and valuable comments and suggestions. This work has been partly supported from ERDF "Research and Development of Intelligent Components of Advanced Technologies for the Pilsen Metropolitan Area (InteCom)" (no.: CZ.02.1.01/0.0/0.0/17\_048/0007267) and by Grant No. SGS-2019-018 Processing of heterogeneous data and its specialized applications.

\clearpage

\appendix
\section{Efficient LP-ICP Approach Based on Minimum Nonconformity Change}
\label{appendix}

In this section we provide the mathematical foundations for the developed process of constructing the set $Q_{n+m}$ for each test instance $x_{n+m}$. First, assumptions, auxiliary definitions and Lemmas are given. Next we provide a process of finding $Q(t_{n+m})$ in Section \ref{construct_set_q_based_on_minimum_nonconformity_change} and a proof that $\Gamma_{n+m}^{\epsilon} \subseteq Q(t_{n+m}) $ in Proposition \ref{proposition_1}. We consider the multi-hot representations for label-sets (see Figure \ref{fig:multi-hot}), i.e. $\vv{Y}_j = (Y_j^1, \dots, Y_j^d )$. 

\subsection{Assumptions}
\label{app:assumptions}
We consider the general efficient LP-ICP framework as described in Section \ref{sec:efficient_lp_icp}. Additionally we assume: 
\begin{itemize}

\item
The nonconformity measure $A \in L_p, p \geq 1$ as in Section \ref{sec:nonconf}. That is :
\begin{subequations}
    \label{eq:nonconformity2}
    \begin{align}
    \alpha_i = A(\vv{o}_i,\vv{y}_i) = \|\vv{o}_i - \vv{y}_i \|_{p} \mspace{35 mu}
    \\
    \alpha^{\vv{Y}_j}_{n+m} = A(\vv{o}_{n+m},\vv{Y}_j) = \|\vv{o}_{n+m} - \vv{Y}_j \|_{p},
    \end{align}
\end{subequations}
where classifier's raw prediction for test instance $x_{n+m}$ is $\vv{o}_{n+m} = (o^1_{n+m},\dots, o^d_{n+m})$, $o^k_{n+m} \in [0,1], k = 1,\dots, d$, the actual label-set (true target) for calibration instance $i$ is $\vv{y}_i$ and possible label-set is $\vv{Y}_j \in \{\vv{Y}_1, \dots, \vv{Y}_c\} \subseteq \mathcal{P}(\{\Psi_1, \dots, \Psi_d\})$. 
\end{itemize}

\subsection{Auxiliary Definitions}
\label{auxiliary_definitions}

We define :

\begin{itemize}

    \item
    The underlying classifier's prediction vector $\vv{z}_{n+m}$ which is the binary vector that derives from thresholding the raw prediction $\vv{o}_{n+m}$ with $0.5.$ i.e.  $\vv{z}_{n+m} = (z^1_{n+m}, \dots, z^d_{n+m})$,  $z^k_{n+m} = \lfloor o^k_{n+m} + 0.5 \rfloor \in \{0, 1\}$ \footnote{The floor function $\lfloor w \rfloor$ is the function that takes as input a real number $w, w \in \mathbb{R}$, and outputs the greatest integer less than or equal to w.}.

    \item 
    The vector transformation $\vv{R} : (\mathbb{R}^d \times \mathbb{N}) \rightarrow \mathbb{R}^d $ of any binary vector $\vv{e} = (e^1, \dots, e^d), e^k \in \{0,1\}, k = 1, \dots, d$ that reverses the $l^{th}$ component (label) of $\vv{e}$, from 0 to 1 or vice verse, i.e. $\vv{R}(\vv{e}, l) = (e^1, \dots,e^{l-1},1- e^l, e^{l+1}, \dots, e^d)$.
    
    \item
    The certainty vector $\vv{u}_{n+m}$ that indicates the certainty level of the underlying classifier's raw prediction $\vv{o}_{n+m}$ for every label $\Psi_k$, i.e. $\vv{u}_{n+m} = (u^1_{n+m},\dots,u^d_{n+m})$, $u^k_{n+m} = |o^k_{n+m} - 0.5|, k = 1, \dots, d$.

    \item
    The index vector $\vv{s}_{n+m} = (s^1_{n+m},\dots,  s^d_{n+m}), s^k_{n+m} \in \{1,\dots,d\}$ for sorting (in ascending order) the labels $\Psi_1, \dots, \Psi_d$ according to the certainty level of underlying classifier's raw predictions, i.e. for the resulting ordering : $\Psi_{s^1_{n+m}}, \dots, \Psi_{s^d_{n+m}}$ and $(\forall f, h \in \{1, \dots, d\})(f \leq h) \implies u^{s^{f}_{n+m}}_{n+m} \leq u^{s^{h}_{n+m}}_{n+m}$.

\end{itemize}

\subsection{Lemmas}

\begin{lemma}
\label{lemma_1}

For test instance $x_{n+m}$, nonconformity measure $A \in L_p, p \geq 1$ and label-sets  $\vv{Y}_j \in \{\vv{Y}_1, \dots, \vv{Y}_c\}$ the following holds true:
$\min\limits_{j = 1, \dots, c}(\alpha^{\vv{Y}_j}_{n+m}) = \alpha^{\vv{z}_{n+m}}_{n+m}$

\end{lemma}

\begin{proof}
$\min\limits_{j = 1, \dots, c}(\alpha^{\vv{Y}_j}_{n+m}) =
\\
\min\limits_{j = 1, \dots, c}\left(\|\vv{o}_{n+m} - \vv{Y}_j \|_{p}\right)= 
\\
\min\limits_{j = 1, \dots, c}
\left( \sqrt[p]{|o^{1}_{n+m} - Y^1_j|^p + \dots + |o^d_{n+m} - Y^d_j|^p}\right) =
\\
\sqrt[p]{\min\limits_{j = 1, \dots, c}
\left( |o^1_{n+m} - Y^1_j|^p + \dots + |o^d_{n+m} - Y^d_j|^p \right)} = \\
\sqrt[p]{\min\limits_{j = 1, \dots, c} \left( |o^1_{n+m} - Y^1_j|^p \right) + \dots + \min\limits_{j = 1, \dots, c} \left( |o^d_{n+m} - Y^d_j|^p \right)} =
\\
\sqrt[p]{|o^1_{n+m} - z^1_{n+m}|^p + \dots + |o^d_{n+m} - z^d_{n+m}|^p} = \alpha^{\vv{z}_{n+m}}_{n+m}$. 
\end{proof}

\begin{lemma}
\label{lemma_2}
For test instance $x_{n+m}$, nonconformity measure $A \in L_p, p \geq 1$, the following holds true :
$u^f_{n+m} \leq u^h_{n+m} \Leftrightarrow$
$\alpha^{\vv{R}(\vv{z}_{n+m},f)}_{n+m} \leq \alpha^{\vv{R}(\vv{z}_{n+m},h)}_{n+m}$  
\end{lemma}

\begin{proof}
$u^f_{n+m} \leq u^h_{n+m} \Leftrightarrow $
\begin{equation}
\label{eq:lemma_2_hypothesis}
|o^f_{n+m} - 0.5| \leq |o^h_{n+m} - 0.5| 
\end{equation}
\\
$\alpha^{\vv{R}(\vv{z}_{n+m},f)}_{n+m} \leq \alpha^{\vv{R}(\vv{z}_{n+m},h)}_{n+m} \Leftrightarrow$
\\
$\sqrt[p]{|o^1_{n+m} - z^1_{n+m}|^p + \dots + |o^f_{n+m} - 1 + z^f_{n+m}|^p + \dots + |o^d_{n+m} - z^d_{n+m}|^p} \leq \\ 
\sqrt[p]{|o^1_{n+m} - z^1_{n+m}|^p + \dots + |o^h_{n+m} - 1 + z^h_{n+m}|^p + \dots + |o^d_{n+m} - z^d_{n+m}|^p}  \Leftrightarrow $
\begin{equation}
\label{eq:lemma_2_objective}
    |o^f_{n+m} - 1 + z^f_{n+m}|^p \leq |o^h_{n+m} - 1 + z^h_{n+m}|^p
\end{equation}

We distinguish all possible cases for $o^f_{n+m}$ and $o^h_{n+m}$ and prove (\ref{eq:lemma_2_hypothesis}) $\Leftrightarrow$ (\ref{eq:lemma_2_objective}):

\begin{itemize}
    \item Case: $o^f_{n+m} \geq 0.5$ and $o^h_{n+m} \geq 0.5$
    \\
    Then $z^f_{n+m} = 1, z^h_{n+m} =1$, and : 
    \\
    (\ref{eq:lemma_2_hypothesis}) $\Leftrightarrow o^f_{n+m}  \leq o^h_{n+m} \Leftrightarrow o^f_{n+m} -1 + z^f_{n+m} \leq o^h_{n+m} -1 +z^h_{n+m} \Leftrightarrow $ (\ref{eq:lemma_2_objective}).
    
    \item Case: $o^f_{n+m} \geq 0.5$ and $o^h_{n+m} < 0.5$
    \\
    Then $z^f_{n+m} = 1, z^h_{n+m} = 0$, and : 
    \\
    (\ref{eq:lemma_2_hypothesis}) $ \Leftrightarrow o^f_{n+m} - 0.5  \leq - o^h_{n+m}  +0.5 \Leftrightarrow
    o^f_{n+m} \leq - o^h_{n+m}  +1 \Leftrightarrow
    |o^f_{n+m} - 1 + 1|^p \leq |o^h_{n+m} - 1 + 0|^p \Leftrightarrow$
    (\ref{eq:lemma_2_objective}).
    
    \item Case: $o^f_{n+m} < 0.5$ and $o^h_{n+m} \geq  0.5$
    \\
    Then $z^f_{n+m} = 0, z^h_{n+m} = 1$, and : 
    \\
    (\ref{eq:lemma_2_hypothesis}) $\Leftrightarrow - o^f_{n+m} + 0.5  \leq  o^h_{n+m} - 0.5 \Leftrightarrow  - o^f_{n+m} +1 \leq  o^h_{n+m}  \Leftrightarrow  |o^f_{n+m} - 1 + 0|^p \leq |o^h_{n+m} - 1 + 1|^p \Leftrightarrow$ 
    (\ref{eq:lemma_2_objective}).

    \item Case: $o^f_{n+m} < 0.5$ and $o^h_{n+m} < 0.5$
    \\
    Then $z^f_{n+m} = 0, z^h_{n+m} =0$, 
    and:
    \\
    (\ref{eq:lemma_2_hypothesis}) $ \Leftrightarrow - o^f_{n+m} +0.5 \leq - o^h_{n+m} +0.5 \Leftrightarrow    - o^f_{n+m} +1 \leq  - o^h_{n+m} +1 \Leftrightarrow    |o^f_{n+m} - 1 + 0|^p \leq |o^h_{n+m} - 1 + 0|^p \Leftrightarrow$ (\ref{eq:lemma_2_objective}).
    \end{itemize}

\end{proof}

\subsection{Intuition Behind Constructing $Q$ based on Minumum Nononconformity Change}
\label{construct_set_q_based_on_minimum_nonconformity_change}
Let set $Q$ be defined for each test instance as $Q(t_{n+m}) = \{\vv{Y}_j: \sum^d_{i=1} Y^i_j \oplus z_{n+m}^i < t_{n+m}\}$, where $\oplus$ is the xor operator and $t_{n+m} \in \{1,\dots,d\}$. In such a setting the implementation of efficient LP-ICP is achieved by finding the minimum  $t_{n+m} = t_{n+m}(\alpha_{i_0}(\epsilon))$ such that $ \Gamma_{n+m}^{\epsilon} \subseteq Q(t_{n+m})$,  that is:

\begin{equation*}
    t_{n+m} = \min\limits_{\nu = 1,\dots,d} (\nu) : \alpha^{ \vv{R}(\dots(\vv{R}(\vv{z}_{n+m},s^{1}_{n+m}),\dots), s^{\nu}_{n+m})}_{n+m} > \alpha_{i_0} 
\end{equation*}

The intuition behind the definition of $t_{n+m}$ is as follows:
\begin{itemize}
\item 
For $\alpha^{\vv{z}_{n+m}}_{n+m} > \alpha_{i_0}$:

Then $\Gamma_{n+m}^{\epsilon} = \emptyset$ which is taken for $t_{n+m} = 0$, because of Lemma \ref{lemma_1}

\item
For $\alpha^{\vv{z}_{n+m}}_{n+m} \leq \alpha_{i_0}$:

Using Lemma \ref{lemma_2} we get: $\alpha^{\vv{R}(\vv{z}_{n+m},s^{1}_{n+m})}_{n+m} \leq \alpha^{\vv{R}(\vv{z}_{n+m},s^{k}_{n+m})}_{n+m}, k = 1,\dots, d $, which means that the minimal nonconformity change of $\alpha^{\vv{z}_{n+m}}_{n+m}$ is achieved for the label-set $\vv{R}(\vv{z}_{n+m},s^{1}_{n+m})$. 

\begin{itemize}
\item
For  $\alpha^{\vv{R}(\vv{z}_{n+m},s^{1}_{n+m})}_{n+m} > \alpha_{i_0}$ then there is no other label-set apart from $\vv{z}_{n+m}$ that can be included in $\Gamma_{n+m}^{\epsilon}$. The objective is taken for $t_{n+m} = 1$.

\item
For $\alpha^{\vv{R}(\vv{z}_{n+m},s^{1}_{n+m})}_{n+m} \leq \alpha_{i_0}$ we reapply the minimal change to the nonconformity score $\alpha^{\vv{R}(\vv{z}_{n+m},s^{1}_{n+m})}_{n+m}$ which is the one corresponding to the label-set $\vv{R}(\vv{R}(\vv{z}_{n+m},s^{1}_{n+m}), s^{2}_{n+m})$.
We repeat this process until $\alpha_{i_0}$ is surpassed. Then the objective is taken for
$t_{n+m} = \nu$, where $\nu$ is the number of times the minimum nonconformity change was applied, i.e. $t_{n+m} = \min\limits_{\nu = 1,\dots,d} (\nu)$ : $\alpha^{ \vv{R}(\dots(\vv{R}(\vv{z}_{n+m},s^{1}_{n+m}),\dots), s^{\nu}_{n+m})}_{n+m} > \alpha_{i_0} $.

\end{itemize}
\end{itemize}

\begin{proposition}[$ \Gamma_{n+m}^{\epsilon} \subseteq Q_{n+m}$]
\label{proposition_1}
For test instance $x_{n+m}$, pre-specified significance level $\epsilon$, nonconformity measure $A \in L_p, p \geq 1$,  and set $Q(t_{n+m}) = \{\vv{Y}_j: \vv{Y}_j: \sum^d_{i=1} Y^i_j \oplus z_{n+m}^i < t_{n+m}\}$ where $t_{n+m} = \min\limits_{\nu = 1,\dots,d} (\nu)$ : $\alpha^{ \vv{R}(\dots(\vv{R}(\vv{z}_{n+m},s^{1}_{n+m}),\dots), s^{\nu}_{n+m})}_{n+m} > \alpha_{i_0} $ the following holds true:
$\forall \vv{Y}_j \notin Q_{n+m} \implies 
\vv{Y}_j \notin \Gamma_{n+m}^{\epsilon}$
\end{proposition}

\begin{proof} 
\qquad $ \vv{Y}_j \notin Q_{n+m} \Leftrightarrow 
\\
\sum^d_{i=1} Y^i_j \oplus z_{n+m}^i \geq t_{n+m} \Leftrightarrow 
\\
\vv{Y}_j = \vv{R}(\dots(\vv{R}(\vv{z}_{n+m},l_1),\dots), l_k), 
\\
\text{where } l_i \in \{1,\dots,d\}, k \geq t_{n+m}$, and $l_i$ differ from each other, placed in ascending order \footnote{i.e. $ (\forall f, h \in \{1,\dots,d\}) (f < h)
\Leftrightarrow (u^{l_f}_{n+m} < u^{l_h}_{n+m})$
} 
\\
\\
By definition of $\vv{s}_{n+m}$:
\\
$u^{l_i}_{n+m} \geq u^{s^i_{n+m}}_{n+m}, \forall i \in \{1,\dots,k\}$
\\
Using $t_{n+m}$ times Lemma \ref{lemma_2}  we get: 
\\
\begin{equation}
\label{equation_to_combine_0}
\alpha^{\vv{R}(\dots(\vv{R}(\vv{z}_{n+m},l_1),\dots), l_{t_{n+m}})}_{n+m} \geq 
\alpha^{\vv{R}(\dots(\vv{R}(\vv{z}_{n+m},s^1_{n+m}),\dots), s^{t_{n+m}}_{n+m})}_{n+m}
\end{equation}
\\
Since $k \geq t_{n+m}$: 
\begin{equation}
\label{equation_to_combine_1}
\alpha^{\vv{R}(\dots(\vv{R}(\vv{z}_{n+m},l_1),\dots), l_k)}_{n+m} \geq 
\alpha^{\vv{R}(\dots(\vv{R}(\vv{z}_{n+m},l_1),\dots), l_{t_{n+m}})}_{n+m} 
\end{equation}
\\
\begin{equation}
(\ref{equation_to_combine_0}) \wedge (\ref{equation_to_combine_1}) \implies
\label{equation_to_combine_2}
\alpha^{\vv{R}(\dots(\vv{R}(\vv{z}_{n+m},l_1),\dots), l_k)}_{n+m} \geq 
\alpha^{\vv{R}(\dots(\vv{R}(\vv{z}_{n+m},s^1_{n+m}),\dots), s^{t_{n+m}}_{n+m})}_{n+m}
\end{equation}
\\
By definition of $t_{n+m}$ :
\begin{equation}
\label{equation_to_combine_3} 
\alpha^{ \vv{R}(\dots(\vv{R}(\vv{z}_{n+m},s^{1}_{n+m}),\dots), s^{t_{n+m}}_{n+m})}_{n+m} > \alpha_{i_0}  
\end{equation}

(\ref{equation_to_combine_2}) $\wedge$ (\ref{equation_to_combine_3}) $\implies 
\alpha^{\vv{R}(\dots(\vv{R}(\vv{z}_{n+m},l_1),\dots), l_k)}_{n+m} > \alpha_{i_0} \Leftrightarrow
\\
\alpha^{\vv{Y}_j}_{n+m} > \alpha_{i_0} \Leftrightarrow
\\
\vv{Y}_j \notin \Gamma_{n+m}^{\epsilon}$
\end{proof}

\end{document}